\documentclass[journal]{IEEEtran}
\usepackage{hyperref}
\usepackage{enumitem}
\usepackage{pifont}
\usepackage{lipsum} 
\usepackage{slashbox} 
\usepackage{multirow}
\usepackage{algorithm2e}
\usepackage{makecell}
\usepackage{colortbl}
\usepackage{xcolor}
\usepackage{fontenc}
\usepackage{siunitx} %
\usepackage{subfig}
\usepackage{graphicx}
\usepackage{booktabs} % For professional looking tables
\usepackage{multirow}
\usepackage{lettrine}

\ifCLASSINFOpdf
\else
\fi

\begin{document}

\title{Towards Unconstrained Palmprint Recognition on Consumer Devices: a Literature Review}

\author{Adrian-S.~Ungureanu,~%~\IEEEmembership{Member,~IEEE,}
        Saqib~Salahuddin,~%~\IEEEmembership{Fellow,~OSA,}
        and~Peter~Corcoran,~\IEEEmembership{Fellow,~IEEE}~% <-this % stops a space
\thanks{Adrian-S. Ungureanu is with the National University of Ireland, Galway, email: a.ungureanu1@nuigalway.ie}
\thanks{Saqib Salahuddin is with the National University of Ireland, email: s.salah-ud-din1@nuigalway.ie}
\thanks{Prof. Peter Corcoran is with the National University of Ireland, Galway, email: peter.corcoran@nuigalway.ie}
}

\maketitle
\begin{abstract}
%Palmprints provide distinctive features that can be exploited for high recognition accuracy even using low-cost acquisition devices. 
As a biometric palmprints have been largely under-utilized, but they offer some advantages over fingerprints and facial biometrics. Recent improvements in imaging capabilities on handheld and wearable consumer devices have re-awakened interest in the use fo palmprints.
The aim of this paper is to provide a comprehensive review of state-of-the-art methods for palmprint recognition including Region of Interest extraction methods, feature extraction approaches and matching algorithms along with overview of available palmprint datasets in order to understand the latest trends and research dynamics in the palmprint recognition field.        
\end{abstract}


\IEEEpeerreviewmaketitle

\section{Introduction}
%There is a need for more privacy and more personal data protection in consumer electronics. This can be achieved with having several layers of authentication, which provide stronger barriers to identity theft \textbf{(cit. about identity theft)}.\\
The last decade has seen the migration of biometric recognition approaches onto mobile devices by using fingerprint \cite{corcoran2015mobile_biometrics_overview}, face \cite{amos2016openface} or iris \cite{thavalengal2015iris} as an alternative to conventional authentication using PIN numbers or patterns. \\
Two-factor authentication, multi-modal and multi-biometrics are all considered to be viable options improving the security of a system, as they considerably increase the spoofing effort for an attacker \cite{roberts2007biom_atack_vectors}. \\%Having several factors of authentication is considered more difficult as it increases the spoofing effort for an attacker \cite{roberts2007biom_atack_vectors}.
Jain \textit{et al.} \cite{jain2004introduction2Biom} evaluate several biometric features and reach the conclusion that there is no ideal biometric. Alongside the previously mentioned features is another biometric which has not received as much attention: the palmprint. %Although there is no ideal such biometric feature, the previously mentioned face, fingerprint and iris have
However, there are several advantages which palmprint recognition can offer regarding their deployment on consumer devices:
\begin{itemize}
	\item The features contained in a palmprint are similar to fingerprints, but cover a much larger surface. For this reason they are generally considered to be more robust than fingerprints \cite{jain2004introduction2Biom}. 
	\item Palmprints are more difficult to spoof than faces, which are public feature, or fingerprints, which leave traces on many smooth surfaces.
	\item There is no extra cost required for acquisition, as long as the device is fitted with a camera (optical sensor) and a flash source (LED or screen).
	\item It has potential for multi-biometric recognition, as it can be used with other hand-based features (fingerprints \cite{genovese2019civemsa}, finger knuckles \cite{meraoumia2011fusion}, wrist \cite{matkowski2019study})
	\item It can be seamlessly integrated into the use case of many consumer devices, such as AR/VR headsets \cite{redrock_palmID}, smartphones \cite{Ungureanu2017}, gesture control systems, driver monitorin systems, etc.
\end{itemize}

The aim of this paper is to provide a comprehensive review focusing on the pipeline of palmprint recognition in order to clarify the current trends and research dynamics in the palmprint recognition based biometric systems. The paper discusses in detail the available datasets of palmprint images and reviews the state-of-the-art methods for palmprint recognition. 

A particular emphasis is placed on the improvement in imaging subsystems on handheld and wearable devices and on recent developments in unconstrained palmprint analysis, including the recent availability of new datasets and Region of Interest (ROI) extraction methodologies. 

The rest of the paper is organized as follows. Section II describes existing datasets of palmprint images. Section III provides an overview of approaches developed for the palmprint ROI extraction from various palmprint datasets. Section IV presents an overview of approaches of feature extraction and matching algorithms. Section V presents discussions and concludes the paper. 

\begin{figure}
	\centering
	\includegraphics[width=\linewidth]{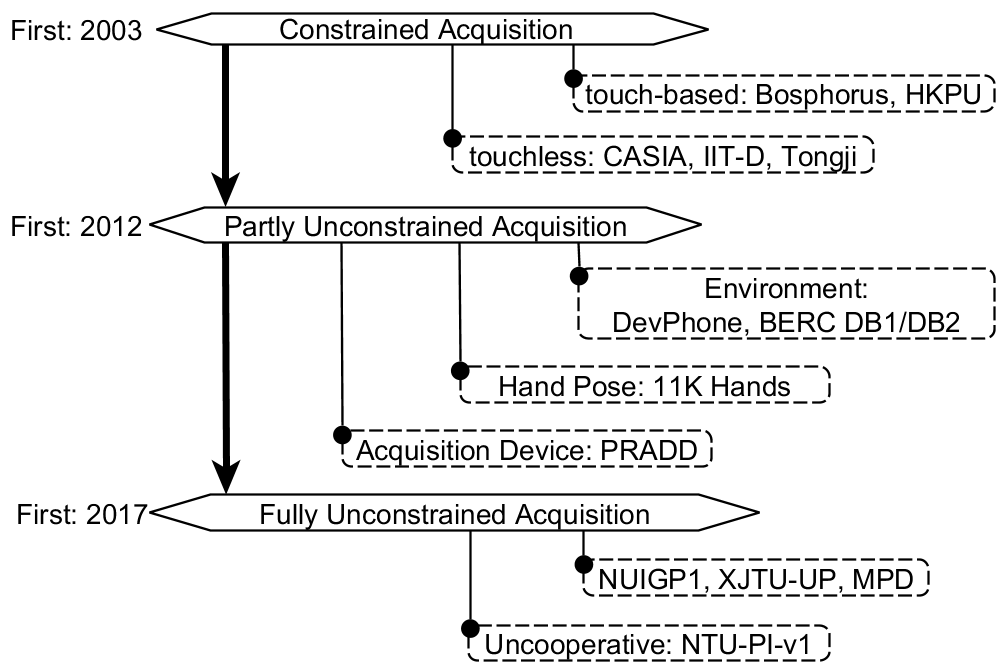}  
	\caption[Overview of current palmprint datasets]{Timeline overview of palmprint datasets, based on how constrained their environment of acquisition is.} 
	\label{palmprint_datasets_overview}
\end{figure}

\section{Palmprint Datasets}
%*******************************************************************************
% detailed presentation on existing Palmprint databases (emphasis on images acquired with smartphones) 
%Introduction to palmprint datasets.
This section presents an overview of palmprint datasets used for the recognition of palmprints in the visible spectrum (hyperspectral imaging at various wavelengths is not considered, nor 3D acquisition).
%Palmprint datasets represent an important part of the resources used by the research community. All datasets have the following characteristics in common:

The currently available palmprint datasets can be split into three categories, based on the restrictions imposed to the user during the acquisition process (as represented in Fig. \ref{palmprint_datasets_overview} and summarized in Table \ref{pp_datasets}):%. These constraints are generally introduced to aid the extraction of palmprint templates from the hand images:
\begin{enumerate}
	\item Constrained acquisition: This category includes the most popular palmprint datasets, which place the main focus on the feature extraction and matching stages, simplifying the acquisition as much as possible (for the recognition system). Images tend to display hands with a specific hand pose (fingers straight and separated) against a uniform background with no texture, usually black.%Both(Constrained hand pose), (Constrained environment)
	\item Partly unconstrained acquisition: 
	\begin{itemize}
		\item \textbf{Unconstrained} environment: The background is unconstrained, which corresponds to the use case of consumer devices. The hand pose is required to follow a specific protocol, generally consisting of presenting the fingers spread out in front of the sensor (preferably the center of the image).%(Constrained hand pose), (\textbf{Unconstrained} environment)
		\item \textbf{Unconstrained} hand pose: Allows the user to choose the pose of the hand during acquisition. This corresponds to the general expectations  for consumer devices, which require a simplified (and intuitive) protocol of interaction.	%(\textbf{Unconstrained} hand pose), (Constrained environment)
		\item \textbf{Multiple devices} used for acquisition: Matching biometric templates across several devices. Generally the other aspects of the acquisition process (hand pose and background) are constrained.%	Choosing to employ several devices for acquisition instead of only one (not to be confused with multi-biometrics), but 
	\end{itemize}
	\item Fully unconstrained acquisition: \textbf{Unconstrained} environment and hand pose, this represents the most unconstrained scenario, where all conditions of acquisition are left to the choice of the user. A further step is closer to forensic recognition, where the acquisition is uncooperative 
	
	A further subcategory would be the acquisition sce%(\textbf{Unconstrained} hand pose) and (\textbf{Unconstrained} environment)
\end{enumerate}

\subsection{Constrained Palmprint Datasets}
%**************************************************************************************************%

%%% Palmprint datasets 
%%% (Constrained hand pose), (Constrained environment)
%\textbf{Introduce Table with touch-based and touch-less palmprint datasets.}
\begin{table*}[h]
	\caption{Constrained palmprint datasets: (A1) touch-based and (A2) touch-less acquisition, having one hand pose (spread fingers). Partly unconstrained palmprint datasets: (B1) unconstrained environment/background, (B2) multiple devices used during acquisition and (B3) unconstrained hand pose. Fully unconstrained palmprint datasets (C1), as close as possible to the realistic deployment of a palmprint recognition system on smartphones (or similar consumer devices) and (C2) reflecting recognition in an uncooperative environment, closer to forensic recognition. 
	} \label{pp_datasets} % pp_datasets_cat1
	\centering
	\scriptsize %\footnotesize
	%\resizebox{\columnwidth}{!}{%
	\begin{tabular}{|p{5pt}||p{15pt}|p{57.5pt}| p{70pt}| p{25pt}|p{22.55pt}|p{230pt}|}
				  %{|p{5pt}||p{15pt}|p{57.5pt}|p{25pt}| p{50pt}| p{25pt}|p{22.55pt}|p{215pt}|}
		\cline{1-7}
					& Year & Name & Acq. Device & Hands & Images & Description \\
		\hline \hline
		\multirow{2}{5pt}[-1em]{\textbf{A1}} & 2003 &\textbf{HKPU} \cite{TheHongKongPolytechnicUniversity} & scanner & 386 & 7,752 &  Cropped hand images, intended for performance comparison. Two sessions. \textbf{Graylevel} images with black background. \\
		
		\cline{2-7}%\hline
		& 2010 &\textbf{Bosphorus} \cite{Bosphorus_hand_database} & scanner  & 1,560 & 4,846 & \textbf{RGB} images of the entire hand. Hand geometry also used.\\
		
		\hline \hline
		\multirow{7}{5pt}[-1em]{\textbf{A2}} & 2005 &\textbf{CASIA}\cite{CASIA} & digital camera & 624 & 5,502 &  \textbf{Graylevel} images, black background. Uniform lighting. Cropped fingers. \\   
		
		\cline{2-7}%\hline
		 & 2006 &\textbf{IIT-D v1} \cite{IITDelhi2014} & digital camera & 470 & 3,290 & \textbf{RGB} images, black background. Uniform lighting. Cropped wrist. \\
		
		\cline{2-7}%\hline
		 & 2010 &\textbf{COEP} \cite{coep_palmprint_dataset} & digital camera & 168 & 1,344 & \textbf{RGB} images, black background. Pegs used for pose/scale control. \\
		\cline{2-7}%\hline
		 & 2011 &\textbf{GPDS-CL1} \cite{ferrer2011GPDS} & 2 webcams & 100 & 2,000 & \textbf{RGB} images, acquired in both visible and 850 nm. Hand geometry also used.\\  
		\cline{2-7}%\hline
		 & 2017 &\textbf{Tongji} \cite{ZHANG2017199} & digital camera & 1,200& 12,000 &  \textbf{RGB} images, black background. Large-scale dataset.\\
		\cline{2-7}%\hline
		 & 2018 &\textbf{PolyU-IITD v3}\cite{kumar2018HKPU_IITD_v3} & 2 digital cameras & 700 & - & \textbf{RGB} images, black background. Contains images from 2 ethnicities. Significant physical variation considered, and long interval acquisition (15 years). \\
		\cline{2-7}%\hline
		 & 2019 & \textbf{HFUT} \cite{xiao2019palm_roi} & digital camera & 800 & 16,000 & \textbf{RGB} images, black background. Whole hands: fingers and wrist.\\
		\hline \hline
		
		\multirow{5}{5pt}[-4em]{\textbf{B1}} & 2013 &\textbf{DevPhone} \cite{Aoyama2013} & 1 smartphone camera & 30 & 600 & Acquisition controlled with a square guide displayed on the screen; no information about the environment of acquisition. \\
		\cline{2-7}%\hline 
		
		 & 2015 & \makecell[l]{\textbf{BERC} \cite{Kim2015} \textbf{DB1}\\ \textbf{DB2}} & 1 smartphone camera & 120 & \makecell[l]{8,957\\9,224} & Unconstrained environment; controlled hand orientation using a visual guide on the device screen; indoor (DB1) and outdoor (DB2) background. \\
		\cline{2-7}%\hline
		
		& 2016 &\textbf{Tiwari} \textit{et al.} \cite{tiwari2016orb_pp}& 1 smartphone camera& 62 & 182 videos & Users positioned the palm in a circular guide displayed on the device's screen. Each user filmed 3 short videos of the palm.\\
		\cline{2-7}%\hline
		& 2019 &\textbf{BMPD} \cite{izadpana2019novel_mobile_datasets} & 1 smartphone camera & 41 & 1,640 & Hand images against black background. The 2nd acquisition session used stronger angles relative to the hands. \\
		\cline{2-7}%\hline
		& 2019 &\textbf{SMPD} \cite{izadpana2019novel_mobile_datasets} & 1 smartphone camera & 110 & 4,400 & Hand images with black background and flashlight turned on. Included 4 scenarios of hand orientation, including tilted away from the camera. \\
		
		\hline \hline
		% Device constraint removed...
		\multirow{2}{5pt}[-1em]{\textbf{B2}} & 2012 &\textbf{Choras} \textit{et al.} \cite{Choras2012} & 3 smartphone cameras & 84 & 252 & \textbf{RGB} hand images with black background.\\ 
		\cline{2-7}%\hline
		& 2012 & \textbf{PRADD} \cite{Jia2012} & 1 digital camera, 2 smartphones & 100 & 12,000 &  Hand images against black background. Two lighting cases considered. Acquisition not performed by user. \\
		
		\hline \hline
		% Hand pose constraint removed...
		\multirow{2}{5pt}[-2em]{\textbf{B3}} & 2017 & \textbf{11k Hands} \cite{afifi201711k} & 1 digital camera & 380 & 11,076 & RGB images of hands (palm and dorsal) against white background. Variable hand pose. \\
		\cline{2-7}%\hline
		& 2019 &\textbf{NUIGP2} \cite{nuigpalmII2020dataset} & 1 webcam & 52 & 24,631 & \textbf{RGB} images, with variable hand pose and scales, several backgrounds. Intended for training of  palmprint ROI extraction algorithms.\\
		\hline \hline
		
		\multirow{4}{5pt}[-1em]{\textbf{C1}}& 2017  &\textbf{NUIGP1} \cite{Ungureanu2017} & 5 smartphone cameras & 81 & 1,816 & Two lighting conditions, two backgrounds.\\ 
		
		\cline{2-7}%\hline
		& 2019  &\textbf{XJTU-UP} \cite{shao2019efficient} & 5 smartphone cameras & 200 & 30,000+ & Unconstrained background, scenario with enabled flashlight included.\\ 
		\cline{2-7}%\hline
		& 2019  &\textbf{MPD} \cite{zhang2019palm_roi} & 2 smartphone cameras & 200 & 16,000  & Images include rotation but appear to have a single hand pose (spread fingers), to facilitate key-point detection.\\ 
		
		\cline{2-7}%\hline
		& 2019 & \textbf{NTU-CP-v1} \cite{matkowski2019palmprint} & 1 digital cameras & 655 & 2,478 & Unconstrained pose, white background.\\
		
		\hline \hline %\cline{2-7}
		\textbf{C2} & 2019 & \textbf{NTU-PI-v1} \cite{matkowski2019palmprint} & many devices & 2,035 & 7,781 &  Collected from the Internet. Large variation in scale, pose and background.\\
		
		% & 2019  & \makecell[l]{\textbf{NTU-CP-v1}\\ \textbf{NTU-PI-v1}} & PP & \makecell[l]{655\\20135} & \makecell[l]{2,478\\7,781} & Uncontrolled and uncooperative environment. \\ 
		\hline
		%fully unconstrained datasets...
	\end{tabular}%}
	\\ Acq. Device = Acquisition Device; Hands = Number of hand classes (some datasets only have images from one hand per participant)
	\\ PP = Palmprint; HG = Hand Geometry
	
\end{table*}

\begin{figure}
	\centering
	\subfloat[NUIG\_Palm1 \cite{Ungureanu2017} \label{nuigp1_image_samples}]{%
		\includegraphics[width=1\columnwidth]{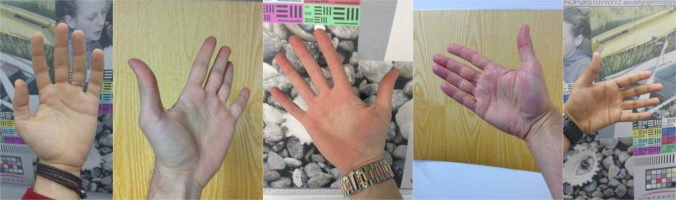}}
	
	%\hfil
	\subfloat[XJTU-UP \cite{shao2019efficient} \label{xjtu_up_samples} source: \textcopyright2019 IEEE]{%
		\includegraphics[width=1\columnwidth]{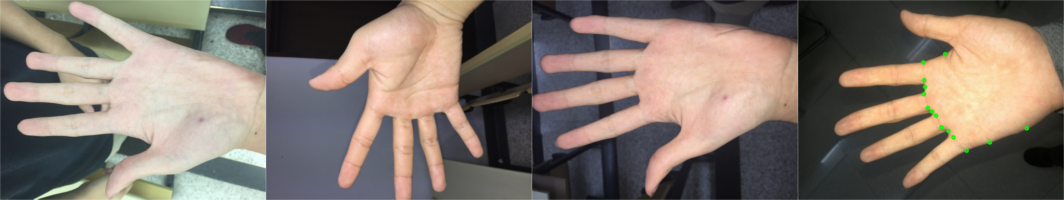}}
	
	%\hfil
	\subfloat[MPD \cite{zhang2019palm_roi} \label{mpd_samples}]{%
		\includegraphics[width=1\columnwidth]{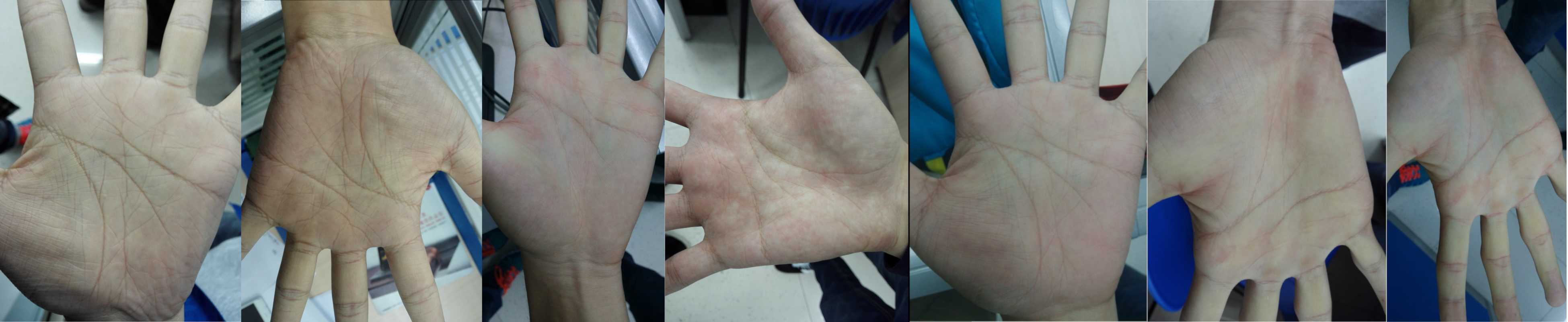}}
	
	\subfloat[NTU-CP-v1 \cite{matkowski2019palmprint} source:  \textcopyright2019 IEEE]{
		\includegraphics[width=1\columnwidth]{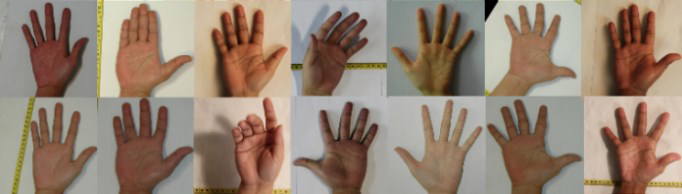}
	}
	
	\subfloat[NTU-PI-v1 \cite{matkowski2019palmprint} source:  \textcopyright2019 IEEE]{
		\includegraphics[width=1\columnwidth]{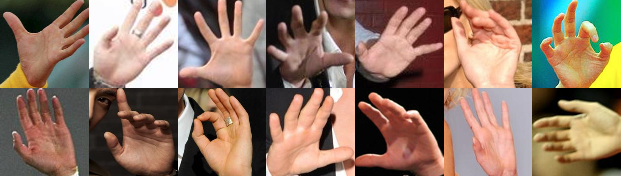}
	}
	\caption[]{Hand image samples from fully unconstrained datasets (C1 and C2) listed in Table \ref{pp_datasets}.}\label{unconstrained_datasets_samples}
\end{figure}

The Hong Kong Polytechnic University Palmprint dataset (HKPU) \cite{TheHongKongPolytechnicUniversity} was the first to provide a large-scale constrained palmprint dataset to compare recognition performance. The images were acquired using a scanner (\textbf{A1} in Table \ref{pp_datasets}) having a cropped guide around the palm, reducing the impact of fingers' position. %Fig. \ref{polyu_samples} shows few samples from HKPU dataset. 
A similar approach for acquiring palmprints but including the entire hand can be found in the Bosphorus Hand dataset \cite{Bosphorus_hand_database}. %Fig. \ref{bosphorus_samples} shows few samples with spread fingers from Bosphorus dataset.\\
The earliest touch-less palmprint datasets (\textbf{A2} in Table \ref{pp_datasets}) were the ones released by the Chinese Academy of Sciences (CASIA) \cite{CASIA} and by the Indian Institute of Technology in Dehli (IIT-D) \cite{IITDelhi2014}. %, of which several samples are provided in Fig. \ref{casia_samples} and \ref{iitd_samples}, respectively.
Both used a digital camera for acquisition in an environment with uniform lighting. The main differences are the scale and color information contained in IIT-D. The hand images in CASIA are gray scale and have cropped fingers.
The College of Engineering Pune (COEP) \cite{coep_palmprint_dataset} released a touch-less dataset of palmprints, but the acquisition relied on pegs to direct the position of fingers relative to the camera. %Fig. \ref{coep_samples} shows few samples from COEP dataset. \\
Another touch-less dataset was released by Las Palmas de Gran Canaria University under the name GPDS \cite{ferrer2011GPDS}. They used two webcams to acquire palmprint images in two sessions. One of the webcams was adapted to acquire NIR images by removing its IR filter and replacing it with an RGB filter. The dataset is split into images acquired in visible range (GPDS-CL1) and in NIR range (GPDS-CL2). %Some samples from GPDS dataset are shown in Fig. \ref{GPDS_samples}. \\ %GPDS100Contactlesshands2Band database
%Another touch-less dataset of palmprints was released from
In 2017, Zhang \textit{et al.} \cite{ZHANG2017199} released a large-scale dataset (12,000 images) of palmprints acquired with a dedicated device containing a digital camera (Tongji). The acquisition environment was dark with a controlled light source illuminating the palm area. \\%(samples in Fig. \ref{tongji_samples}). \\
%All the palmprint images from datasets in Table \ref{pp_datasets} display hands with open palms and spread fingers.
Recently, Kumar \cite{kumar2018HKPU_IITD_v3} released a large-scale dataset of palmprints entitled PolyU-IITD Contactless Palmprint Database v3, introducing a variety of challenges. Firstly, it contains hand images from two ethnicities (Chinese and Indian). Secondly, the palmprints were acquired from both rural and urban areas. The physical appearance of the hands varies significantly, there being instances of birth defects, cuts and bruises, callouses from manual labour, ink stains and writing, jewelry and henna designs. %as shown in Fig. \ref{polyu_iitd_v3_samples}. 
The dataset also contains a 2nd acquisition session after 15 years, for 35 subjects.

% FIGURE: PALMPRINT SAMPLES FROM CONSTRAINED DATASETS
%\begin{figure*}
%	\centering
%	\subfloat[HKPU \cite{TheHongKongPolytechnicUniversity} \label{polyu_samples}]{%
%		\includegraphics[width=.625\textwidth]{images/constrained_dbs/hkpu_samples}}
%	\hfil
%	%\unskip\ \vrule
%	\subfloat[Bosphorus \cite{Bosphorus_hand_database} \label{bosphorus_samples}]{%
%		\includegraphics[width=.375\textwidth]{images/constrained_dbs/bosphorus_samples}}
%	\hfil
%	\subfloat[CASIA \cite{CASIA} \label{casia_samples}]{%
%		\includegraphics[width=.5\textwidth]{images/constrained_dbs/casia_samples}}
%	\hfil
%	\subfloat[IITD \cite{IITDelhi2014} \label{iitd_samples}]{%
%		\includegraphics[width=.5\textwidth]{images/constrained_dbs/iitd_samples}}
%	\hfil
%	\subfloat[COEP \cite{coep_palmprint_dataset} \label{coep_samples}]{%
%		\includegraphics[width=.527\textwidth]{images/constrained_dbs/coep_samples}}
%	\hfil
%	\subfloat[GPDS \cite{ferrer2011GPDS} \label{GPDS_samples}]{%
%		\includegraphics[width=.473\textwidth]{images/constrained_dbs/gpds_samples}}
%	\hfil
%	\subfloat[Tongji \cite{ZHANG2017199} \label{tongji_samples}]{%
%		\includegraphics[width=.5\textwidth]{images/constrained_dbs/tongji_samples}}
%	\hfil
%	\subfloat[PolyU-IITD v3 \cite{kumar2018HKPU_IITD_v3} \label{polyu_iitd_v3_samples}]{%
%		\includegraphics[width=.5\textwidth]{images/constrained_dbs/polyu_iitd_v3_samples}}
%	
%	\setlength{\belowcaptionskip}{-10pt}
%	\caption[]{Hand image samples from constrained datasets (A1, A2) in Table \ref{pp_datasets}.}\label{constrained_datasets_samples} %and Table \ref{pp_datasets_cat2}.
%\end{figure*}	

\subsection{Partly Unconstrained Palmprint Datasets}
%**************************************************************************************************%
Moving away from constrained scenarios, several datasets introduced at least one challenging factor in the context of palmprint recognition systems.% An overview is presented in Table \ref{pp_datasets_cat2}.

% unconstrained Environment/Background
Considering an unconstrained environment for acquisition (\textbf{B1} in Table \ref{pp_datasets}) leads to both variable background and lighting conditions. An initial step was made for palmprint matching in the context of smartphones by Aoyama \textit{et al.} \cite{Aoyama2013} in 2013 with a small dataset of images (called DevPhone). Unfortunately, the conditions of acquisition are not clear (how many backgrounds considered, if flashlight was enabled), besides the fact that users were required to use a square guide to align the palm with the center of the acquired image. % as shown in Fig. \ref{devphone_samples}. \\
A much larger dataset was acquired by Kim \textit{et al.} \cite{Kim2015} both in-doors and out-doors (BERC DB1 and DB2). Both DB1 and DB2 included a scenario where the smartphone's flashlight was enabled. As in the case of DevPhone, the images in BERC DB1/DB2 contained hands with specific hand pose (open palm with spread fingers.\\ %, as can be seen in Fig. \ref{berc_samples}). \\
A different approach to acquisition was provided by Tiwari \textit{et al.} \cite{tiwari2016orb_pp} who recorded videos of palmprints with a smartphone, with the video centered on the user's palmprint. \\
Recently, Izadpanahkakhk \textit{et al.} \cite{izadpana2019novel_mobile_datasets} introduced two palmprint datasets acquired with a smartphone camera - Birjand University Mobile Palmprint Database (BMPD) and Sapienza University Mobnile Palmprint Database (SMPD). The variation considered for investigation was the rotation of the hands (in both datasets), both in-plane and out-of-plane rotation.\\
% Several devices used for acquisition

The first dataset of palmprints acquired with multiple devices (\textbf{B2} in Table \ref{pp_datasets}), albeit of reduced size, was developed by Choras \textit{et al.} \cite{Choras2012} using three smartphones. \\% Several samples are provided in Fig. \ref{choras_samples}.\\
Jia \textit{et al.} \cite{Jia2012} developed a large dataset of images entitled Palmprint Recognition Accross Different Devices (PRADD) using two smartphones and one digital camera. The background used was a black cloth. The hand's posture was restricted. From the images provided in \cite{Jia2012}, it appears that the acquisition was performed by someone other than the participants. % Several images from each device class are presented in Fig. \ref{pradd_samples}. \\
Unfortunately, the datasets developed by Choras \textit{et al.} \cite{Choras2012} and Wei \textit{et al.} \cite{Wei2016} are currently not available to the research community.

% Unconstrained hand pose 
The first palmprint dataset to consider the hand pose variation (\textbf{B3} in Table \ref{pp_datasets}), understood as open palms with spread fingers versus closed fingers, was collected by Afifi \textit{et al.} and released under the name 11K Hands \cite{afifi201711k}. It contains over 11,000 images of hand images - both palmar and dorsal (each has about 5,500 images). The images were acquired against a white background, using a digital camera. %Some samples from 11K Hands are provided in Fig. \ref{11khands_samples}. 
An auxiliary palmprint dataset exploring various hand poses was released in 2019 by the authors under the name NUIG\_Palm2 (NUIGP2) \cite{nuigpalmII2020dataset}. NUIGP2 was designed to support the development of ROI extraction algorithms. %Several samples are presented in Fig. \ref{nuigp2_samples}.

% FIGURE WITH PALMPRINT SAMPLES FROM PARTLY CONSTRAINED DATASETS
%\begin{figure*}
%	\centering
%	\subfloat[DevPhone \cite{Aoyama2013} \label{devphone_samples}]{%
%		\includegraphics[width=.575\textwidth]{images/less_constrained/devphone_samples}}
%	\hfil
%	\subfloat[BERC \cite{Kim2015} \label{berc_samples}]{%
%		\includegraphics[width=.425\textwidth]{images/less_constrained/berc_samples}}
%	
%	\hfil
%	\subfloat[\cite{Choras2012} \label{choras_samples}]{%
%		\includegraphics[width=.36\textwidth]{images//less_constrained/choras_samples}}
%	\hfil
%	\subfloat[PRADD \cite{Jia2012} \label{pradd_samples}]{%
%		\includegraphics[width=.64\textwidth]{images/less_constrained/pradd_samples}}
%	
%	
%	\hfil
%	\subfloat[11K Hands \cite{afifi201711k} \label{11khands_samples}]{%
%		\includegraphics[width=.5\textwidth]{images/less_constrained/11khands_sample}}
%	\hfil
%	\subfloat[NUIG\_Palm2 \label{nuigp2_samples}]{%
%		\includegraphics[width=.5\textwidth]{images/less_constrained/nuigp2_samples}}
%	
%	%\setlength{\belowcaptionskip}{-10pt}
%	\caption[]{Hand image samples from partly unconstrained datasets listed in Table \ref{pp_datasets_cat2}.}\label{less_constrained_datasets_samples}
%\end{figure*}

%**************************************************************************************************%
\subsection{Fully Unconstrained Palmprint Datasets}
%**************************************************************************************************%
%%% (Unconstrained hand pose), (Unconstrained environment)
% NUIG_Palm1, XJTU-UP
% Table3, with description of these datasets	
% describe CATEGORY C2! Add some pictures from those 2 new datasets...

This category of palmprint datasets attempts to bring to researchers conditions as close as possible to a realistic deployment of a palmprint recognition system on consumer devices. An overview is presented in Table \ref{pp_datasets} for categories \textbf{C1} and \textbf{C2}.

The first dataset to provide such palmprint images was released in 2017 by Ungureanu \textit{et al.} \cite{Ungureanu2017} under the name NUIG\_Palm1 (NUIGP1). It contains images from several devices in unconstrained scenarios (both background and hand pose, as presented in Fig. \ref{nuigp1_image_samples}). \\
Recently a large-scale dataset of palmprint images acquired in similar conditions to NUIGP1 was released by Shao \textit{et al.}, entitled Xian Jiaotong University Unconstrained Palmprint database (XJTU-UP) \cite{shao2019efficient}. The dataset contains 30,000+ images (200 hands) using five smartphones, making it the largest currently available palmprint dataset acquired with smartphone cameras. Several samples are provided in Fig. \ref{xjtu_up_samples}.\\
Another large-scale palmprint dataset acquired with smartphones was released recently by Zhang \textit{et al} \cite{zhang2019palm_roi}. They used two smartphones to collect 16,000 hand images in unconstrained conditions.

Representing the next step of this trend, the NTU-Palmprints from Internet (NTU-PI-v1) \cite{matkowski2019palmprint} was released in late 2019, where severe distortions in the hand pose represent the main challenge to palmprint recognition. The dataset is especially large in terms of the number of hand classes (2,035), with a total of 7,781 images. Matkowski \textit{et al.} \cite{matkowski2019palmprint} also release a dataset of more conventional hand images where the hand pose varies significantly, with acquisition against white background. This dataset, entitled 'NTU-Contactless Palmprint Database' (NTU-CP-v1) also contains a relatively large number of hand classes (655), with 2,478 hand images in total.

% on smartphones!!!! -> starting with the lit. review from TCE
\section{ROI Template Detection and Extraction}\label{lit_rev_rois} 
% provide definition/stages of the ROI extraction? where should this be inserted?...
This section presents a general overview of existing approaches for palmprint ROI extraction. \\
The process of ROI extraction is an essential part of the palmprint recognition system, as any inconsistencies in ROI templates will affect the recognition task. 

\begin{figure}
	\centering
	\includegraphics[width=1.\columnwidth]{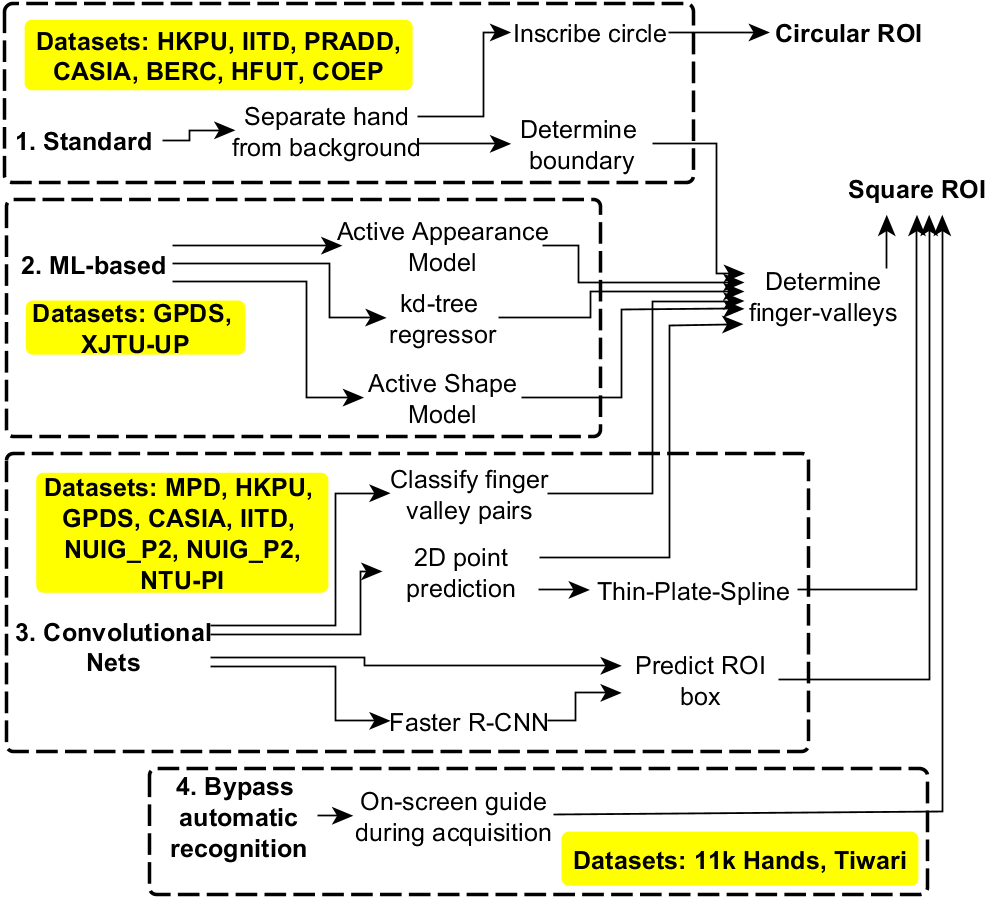}
	\caption[Overview of palmprint ROI extraction approaches]{Overview of approaches for palmprint ROI extraction, with four  categories based on how constrained the datasets are.} \label{roi_extraction_rev_overview}
\end{figure}

The existing ROI extraction techniques can be grouped in four categories, based on the cues contained in the hand images as shown in Fig. \ref{roi_extraction_rev_overview}:
\begin{itemize}
	% Segmentation based approaches: Zhang, etc.
	\item Standard palmprint ROI extraction: algorithms based on separating the hand from the background (segmentation) and performing measurements to determine the landmarks (or palm region) required for ROI extraction. This family of techniques relies on accurate segmentation, as well as a specific hand pose (open palm with spread fingers).
	
	% Machine learning based approaches: Kd trees (Shao), AAM and ASM, etc.
	\item ROI extraction based on conventional Machine Learning (ML) algorithms: ML approaches are used for the detection of palmprints or used for key-point regression. The key-point regression is method that takes a hand image as an input and returns a set of points used for ROI extraction.
	
	% DNN approaches: Bao, Izadpana, 
	\item ROI extraction based on Deep Neural Networks (DNNs):	Approaches relying on DNN soutions to perform detection or key-point regression task.
	
	%\item Corner-based approaches, relying on Harris corner detection \cite{harris1988combined}.
	
	% Holistic approaches, bypassing ROI extraction: Afifi and Tiwari
	\item Avoiding ROI detection altogether: based on specific acquisition protocols.
\end{itemize}

\subsection{Standard Palmprint ROI Extraction}  \label{roi_cat_1}
% will probably be the longest subsection... but should not waste too much time/effort on this...
Standard palmprint ROI extraction algorithms rely on accurate segmentation  of the hand region from the background. The most used approaches include using Otsu's thresholding method \cite{otsu1979threshold} applied to grayscale images, or using a skin-color model \cite{leng2014palm_roi}. The segmentation is a pre-processing stage that characterizes the shape of the hand and determines the key-points required for ROI extraction.

%% Figure outlying the Overview of ROI extraction approaches. Could also be replaced by a Table

The most popular ROI extraction approach was introduced by Zhang \textit{et al.} \cite{zhang2003online} in 2003, which relies on the constrained environment from images in databases (A1, A2) in Table \ref{pp_datasets}, either touch-based or touch-less. Zhang \textit{et al.} ROI extraction approach relies on determining the tangent line between the two side finger valleys in order to normalize the palmprint's rotation and provide a reference point from which to extract a square region. This step is made possible thanks to the constrained environment of acquisition (black background, constant lighting), characteristic of palmprint datasets (A1, A2) in Table \ref{pp_datasets}. \\
Recently, Xiao \textit{et al.} \cite{xiao2019palm_roi} proposed an approach based on the intersection of the binarized hand with lines of specific orientations, resulting in several candidate points for the finger valleys. They then used K-means clustering to obtain the center of each cluster.

A second category of approaches defines the contour of the extracted hand, and the distance from a point of reference (the geometric center \cite{kumar2018HKPU_IITD_v3, zhou2011palm_roi} or the wrist \cite{hao2008palm_roi}, etc) to the pixels found on the contour \cite{Aoyama2013,badrinath2012palm_roi, tiwari2013palm_roi, hammami2014palm_roi, ito2015palm_roi, poinsot2009palm_roi,chen2007palm_roi,charfi2016sift_svm_fusion}. Considering this distribution of distances, the peaks generally correspond to the tips of the fingers, while the local minimas correspond to the finger valleys. These type of approaches are extremely sensitive to segmentation artifacts and generally apply smoothing to the distribution of distances.

A third category traverses all the contour pixels and counts the pixels belonging to the hand region (a circle was considered for sampling). Balwant \textit{et al.} \cite{balwant2015online} introduced specific rules to determine the finger valleys and finger tips, followed by the correct selection of finger valley points that form an isosceles triangle. Goh Kah Ong \textit{et al.} \cite{gohkahong2008palm_roi} considered sampling with fewer points using 3 stages corresponding to circles with greater radius. The outliers resulting from segmentation artifacts were removed with specific rules. Franzgrote \textit{et al.} \cite{Franzgrote2011} further developed the approach proposed by Goh Kah Ong \textit{et al.} by classifying the angles of remaining lines in order to provide a rough rotation normalization step. The finger valley points were then determined with a horizontal/vertical line (depending on the orientation of the hand), having 8 points of transition from non-hand region to hand region.\\
Morales \textit{et al.} \cite{morales2012palm_roi} fitted a circle inside the binarized hand, with its center found equidistantly from the finger valleys (previously determined with the center-to-contour distances).

A fourth category uses the convex hull to describe the concavity of the binarized hand map and finger valleys \cite{chai2016palm_roi,sun2017palm_roi}.

The following are methods that are hard to classify into one category or another, as they either employ very different or combine several of the previously mentioned approaches together.\\ % at different stages. \\
Khan \textit{et al.} \cite{khan2011contourcode} determined the finger tips and the start of the palm by counting the hand-region pixels along the columns. After determining the pixels corresponding to finger valleys, several 2nd order polynomials were used to extrapolate the middle of the finger valleys. The palm's width was used to determine the size of the ROI (70\% of palm size). This approach requires specific hand pose, with hands always rotated towards the left with spread fingers.\\
Han \textit{et al.} \cite{han2007palm_roi} successively cropped the binarized hand image regions corresponding to fingers (after rotation normalization with PCA) by determining the number of transitions from background to hand area. Leng \textit{et al.} \cite{leng2014palm_roi} determined the finger valleys by computing differential maps upward, to the right and the left. The AND operator was applied on these maps, resulting in 4 regions corresponding to all finger valleys. Ito \textit{et al.} \cite{ito2015palm_roi} considered an approach based on line detection after determining the binarized hand region, and subtracting the major lines corresponding to finger edges. Then a distance was computed from center of the palm, allowing the detection of finger valleys even with closed fingers (not relying on spread fingers). Ito \textit{et al.} compared the effectiveness of their approach with three other algorithms \cite{leng2014palm_roi, zhang2003online,  han2007palm_roi}. \\
Liang \textit{et al.} \cite{liang2019palm_roi} used an ROI extraction approach loosely based on \cite{zhang2003online} and \cite{yoruk2006hand_shape_recog}, where the tip of the middle finger was determined and then extended to the center of the palm 1.2 times. This point was then used as a reference to determine the distance to all contour points, allowing the detection of both finger valleys and tips.\\
Wei \textit{et al.} \cite{Jia2012} exploited the constrained nature of acquisition (hand position pose, scale and rotation) to base the ROI extraction on the accurate detection of the heart line's intersection with the edge of the hand (using the MFRAT defined in \cite{jia2008rloc}), performing specific pixel operations to decide on the ROI's center and size. \\
Kim \textit{et al.} \cite{Kim2015} combined several elements for ROI extraction, such as the use of a distance based on a YCbCr model, a specific hand pose (fingers spread) indicated by a guide displayed during acquisition, as well as validating finger valley points by sampling 10 pixels from the determined hand region. \\
Shang \textit{et al.} \cite{shang2012pp_harris_roi} modified the original Harris corner detection algorithm \cite{harris1988combined} in order to locate the points at the middle of finger valleys. However, this approach relied on constrained acquisition, as the background was not overly complex. Another approach using Harris corners was proposed by Javidnia \textit{et al.} \cite{javidnia2015istas}. After obtaining an initial candidate for the hand region based on skin segmentation, the palm region was located using an iterative process based on the strength of the Harris corners.  

However, none of the standard approaches for palmprint ROI extraction can be used in circumstances where the background's color remotely resembles skin color or the hand's pose is not constrained (such as the (C1, C2) datasets in Table \ref{pp_datasets}). Furthermore, one can point out the limitation of skin color segmentation regardless of the chosen color space, based on the inherent inability of classifying a pixel into skin or non-skin \cite{Albiol2001}.

\subsection{Palmprint ROI Extraction based on Conventional ML Algorithms}  \label{roi_cat_2}
% important but short...
There are few approaches using ML algorithms for ROI extraction regressing either a predefined shape or a set of points.\\
Initially, Doublet \textit{et al.} \cite{doublet2006contact} considered to fit an Active Shape Model (ASM) to a number of points describing the shape of a hand (with spread fingers). The model regressed the output of a skin segmentation step, after which the centers of the two finger valleys were used to normalize the hand's rotation. Ferrer \textit{et al.} \cite{ferrer2011GPDS} used a similar ASM to extract the hand from the background in the GPDS-CL1 dataset.\\
Aykut \textit{et al.} \cite{Aykut2015} considered an Active Appearance Model (AAM), which also considered the texture information from the hand's surface. They also provided the first evaluation of predicted key-points. Because the acquisition of images was performed in a considerably constrained environment, no normalization was required relative to the palmprint's scale. Aykut \textit{et al.} preferred to report the error in terms of pixels (from the ground truth points).

%Another method of determining the finger valley key-points is to fit the hand's shape to a predefined model, either with an Active Shape Model, as described by Doublet \textit{et al.} \cite{doublet2006contact}, or with an Active Appearance Model, as described by Aykut \textit{et al.} \cite{Aykut2015}. However, both of these approaches end up restricting the user's hand pose, which affects usability. \\
Recently, Shao \textit{et al.} \cite{shao2019efficient} employed a complex pipeline for ROI extraction for unconstrained palmprint recognition. The approach included an initial stage of palmprint region detection using Histogram of Oriented Gradients (HOG) and a sliding window providing candidate regions at several scales to a pre-trained SVM classifier for palmprint detection. A tree regressor \cite{kazemi2014kd_tree_regressor} (initially developed for face key-point detection) was then used for the landmark regression task applied to all 14 key-points. Unfortunately, Shao \textit{et al.} did not provide details regarding the performance of their ROI extraction, how its accuracy influences the recognition task, or any comparison with prior algorithms.

\subsection{Palmprint ROI Extraction based on Neural Networks}  \label{roi_cat_3}
%**************************************************************************************************%
There have been only a handful of attempts to use Convolutional Neural Networks (CNNs) for the ROI extraction, and most have consisted solely on experimenting on gray-level images. Bao \textit{et al.} \cite{Bao2017} used the CASIA palmprint database \cite{CASIA} to determine the positions of a hand's finger valley points. They used a shallow network composed of 4 Convolutional and 2 Fully-Connected layers, including several Dropout and  MaxPooling layers. The CNN architecture achieved results comparable to Zhang \textit{et al.} \cite{zhang2003online} in stable conditions, but surpassed it when noise was added. Since, a CNN can adapt to noisy or blurred images, the pixel-based approach used by Zhang \textit{et al.} is vulnerable to any kind of image quality degradation.

Izadpanahkakhk \textit{et al.}  \cite{Izadpanahkakhk2018} trained a similar shallow network based on an existing model proposed by Chatfield \textit{et al.} \cite{chatfield2014return}. The network determined a point in the hand image and the corresponding width/height of the palmprint ROI. The network was composed of 5 Convolutional and 2 Fully-connected layers, including several MaxPooling layers and one Local Response Normalization Layer (LRN). The reported results are good for constrained images from HKPU \cite{TheHongKongPolytechnicUniversity}, but the case of in-plane rotated hands was not considered.  

Jaswal \textit{et al.} \cite{jaswal2018deeppalm_roi} trained a Faster R-CNN \cite{fasterrcnn2015faster} model based on Resnet-50 (87 layers) on three palmprint datasets (HKPU, CASIA and GPDS-CL1). They reported lower Accuracy and Recall rates for CASIA  (up to 5\% less) than for HKPU and GPDS-CL1. This can be explained by slightly larger variation in rotation. Similar to \cite{Izadpanahkakhk2018}, the predicted bounding boxes (considered as ROIs) do not include measures for rotation normalization, which considerably affects the recognition rate for the scenario using images from CASIA, as they contain significant rotation variation. Comparatively, images from HKPU and GPDS-CL1 are already normalized rotation-wise.  %They implement a pipeline first classifying the input image into one of the three datasets, followed by the ROI extraction task. The extracted ROIs don't appear to include any compensation for hand rotation, as the predicted bounding boxes are vertically aligned with the image.

Recently, Liu \textit{et al.} \cite{liu2020shifted_triplet_loss} also considered a Fast R-CNN \cite{girshick2015fastRCNN} for palmprint ROI detection. They acquired several videos of palmprints in 11 environments (no other details provided) where the hand pose was varied (from spread to closed fingers, with several hand orientations). These acquisition sessions resulted in 30,000 images that were used for training and testing. For evaluation, Liu \textit{et al.} only considered the percentage of images above a given threshold for Intersection over Union (IoU). However, several important aspects were not covered in Liu \textit{et al.} work: the number of subjects in the training set, the ROI being aligned with the hand (it is maintained vertical regardless of the hand's orientation) or how much an ROI having 60\% IoU (with the ground truth) affects the recognition task.

% add the paper by matkowsky
% review
An especially promising approach was proposed by Matkowski \textit{et al.}, who integrated a Spatial Transformer Network (STN) into ROI-LAnet, an architecture performing the palmprint ROI extraction. The STN was initially proposed by Jaderberg \textit{et al.} \cite{jaderberg2015spatialTransf} to improve the recogniton of distorted digits. This is achieved by learning a thin plane spline transform based on a collection of points, a Grid generator and a bilinear sampler. The STN learns a transformation $T_{\theta}$ that is differentiable with respect to the predicted coordinates $ \hat{\theta} $ based on the input feature map. \\
ROI-LAnet uses a feature extraction network (based on the first 3 MaxPooling stages from the VGG16 network \cite{VGG16_initial}) to obtain the feature map, followed by a regression network providing estimates for the 9 points used for describing the palmprint region (trained initially using L2 loss). The output of ROI-LAnet is a palmprint ROI of fixed size, which is normalized w.r.t. the hand's pose. The authors then include ROI-LAnet into a larger architecture to train it end-to-end using Softmax for loss function.

%Matkowski \textit{et al.} \cite{matkowski2019palmprint} introduce ROI-LAnet, which is an architecture comprised of a feature extraction network (based on the first 3 MaxPooling stages from the VGG16 network \textbf{(citation!!)}), a regression network providing estimates for the 9 points used for describing the palmprints ($\theta$), followed by a Grid generator and a bilinear sampler. The last three elements define a Spatial Transformer Network (SPT) \cite{jaderberg2015spatialTransf}, which learns a transformation $T_{\theta}$ that is differentiable with respect to the predicted coordinates $ \hat{\theta} $. The output of this network is a palmprint ROI of fixed size, which is normalized w.r.t. the hand's pose. 

\subsection{Avoiding the ROI Detection Altogether} \label{roi_cat_4}

%Furthermore, the option of not integrating a rotation normalization step can only be considered when the feature extraction and/or matching phase use features that are inherently rotation invariant (e.g. Scale Invariant Feature Transform (SIFT) \cite{Zhao2013}, modified SIFT \cite{Morales2011} or Oriented FAST and Rotated BRIEF (ORB) \cite{Tiwari2016}). However, the palmprint region still needs to be determined.
Tiwari \textit{et al.} \cite{tiwari2016orb_pp} provided a guide on the screen of the smartphone during acquisition, avoiding the need for an ROI step. Tiwari then used an algorithm to determine the best frames for feature extraction. Similar to Tiwari's approach, Leng \textit{et al.} \cite{leng2018palm_roi} presented a guide on the smartphone's screen, indicating a specific hand pose and orientation for the hand.

Afifi \textit{et al.} \cite{afifi201711k} considered a different approach, having the entire image as the input to a CNN, thus removing any need for an ROI extraction phase. This approach is only feasible because all other parameters in the acquisition environment (background, lighting and hand orientation/scale) are not constant.

%**************************************************************************************************%
\section{Palmprint Feature Extraction and Matching} \label{lit_rev_feat}	%	%	{Standard feature extraction - a literature review}
%**************************************************************************************************%
%\textbf{A good overview is provided by Genovese \textit{et al} in \cite{genovese2019palmnet} (nice Journal paper) and by Dexing \textit{et al} in \cite{zhong2019palmprint_review}.}
\begin{figure}
	\centering
	\includegraphics[width=\linewidth]{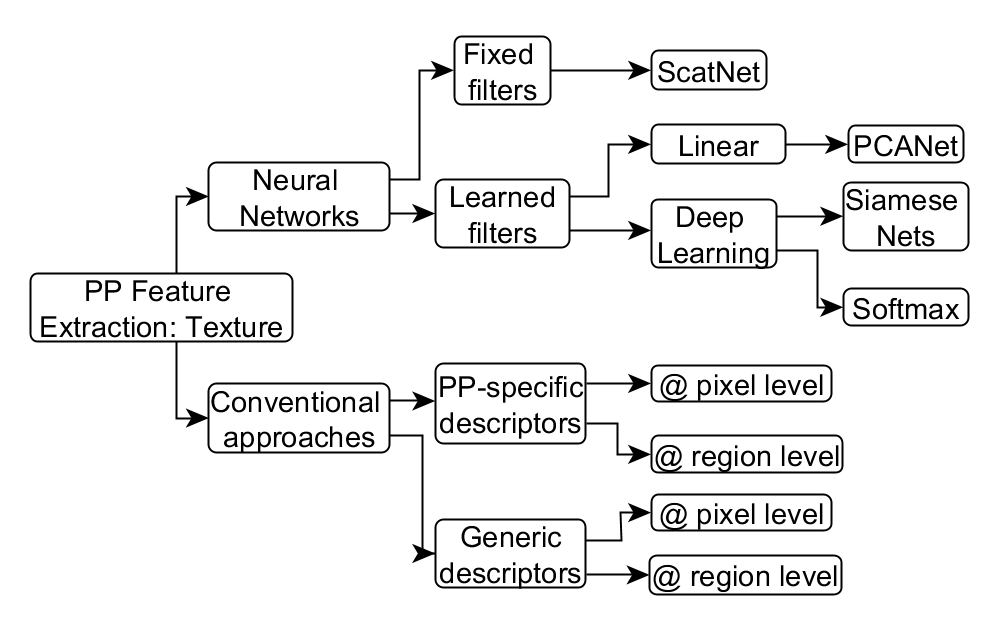} 
	\caption[Overview of palmprint (PP) feature extraction]{Overview of palmprint feature extraction techniques.}\label{palmprint_feature_extract_overview}
\end{figure}

This section presents a general overview of approaches used for palmprint feature extraction, with emphasis being placed on the more recent advancements. 
In this section, the algorithms are split into two categories, based on how the kernels used for feature extraction were obtained (as visualized in Fig. \ref{palmprint_feature_extract_overview}):
\begin{enumerate}
	\item Conventional approaches:
	\begin{enumerate}
		\item Encoding the line orientation \textbf{at pixel-level} with:
		\begin{enumerate}
			\item Generic texture descriptors
			\item Palmprint-specific descriptors.
		\end{enumerate}
		\item Encoding the line orientation \textbf{at region-level}, with:
		\begin{enumerate}
			\item Generic texture descriptors, a special category including descriptors such as SIFT, SURF and ORB, which are treated separately
			\item Palmprint-specific descriptors.
		\end{enumerate}			
	\end{enumerate}
	\item Neural Networks approaches:
	\begin{enumerate}
		\item Having fixed kernels, such as ScatNet \cite{minaee2016palmprint}
		\item Kernels learned based on a training distribution:
		\begin{enumerate}
			\item With no non-linearities, such as PCANet \cite{meraoumia2016palm_pcanet}
			\item Deep Learning approaches:
			\begin{enumerate}
				\item Classifying with Softmax
				\item Using Siamese network architectures.
			\end{enumerate}
		\end{enumerate}
	\end{enumerate}
\end{enumerate}
An overview of the more conventional approaches to palmprint feature extraction is presented in Table \ref{feat_overview_conventional}, whereas an overview of the more recent approaches based on Neural Networks is presented in Table \ref{feat_overview_nn}.

\subsection{Palmprint Feature Extraction - Conventional Approaches}
Conventional palmprint  recognition  approaches  are  mainly  focused on line-like feature detection, subspace learning or texture-based coding. Of these, the best performing approaches have been the texture-based ones \cite{zhang2012palm_review}, which will represent the main focus of this overview. For a broader description of the other groups, please refer to the work of Zhang \textit{et al.} \cite{zhang2012palm_review}, Kong \textit{et al.} \cite{kong2009palmprint_survey} and Dewangan \textit{et al.} \cite{dewangan2012palm_survey}.

Jia \textit{et al.} \cite{jia2017complete_code} defined a framework that generalized the palmprint recognition approaches.% This framework is called Complete Directional Representation and is presented in Fig. \ref{complete_direction_representation_framework}.
The stages of feature encoding are broken down and populated with various approaches. The following sub-sections describe these approaches and provide results in the form of either Equal Error Rate (EER) or Recognition Rate (RR) corresponding to popular palmprint datasets such as HKPU \cite{TheHongKongPolytechnicUniversity}, CASIA \cite{CASIA} or IITD \cite{IITDelhi2014}.

%\begin{figure}[h]
%	\centering
%	\includegraphics[width=\linewidth]{images/complete_direction_representation_framework_jia_2017} 
%	\caption[CDR Framework]{Overview of the framework proposed by Jia \textit{et al} \cite{jia2017complete_code} regarding palmprint feature extraction.} % Overview of the Complete Directional Representation (CDR), as proposed by Jia \textit{et al} 
%	\label{complete_direction_representation_framework}
%\end{figure}

\subsubsection{Extracting Palmprint Features with Texture Descriptors}

Chen \textit{et al.} \cite{chen2010SAX} used a 2D Symbolic Aggregate approximation (SAX) for palmprint recognition. The SAX represents a real valued data sequence using a string of discrete symbols or characters. Applied to grayscale images, it encodes the pixel values, essentially performing a form of compression. The low complexity and high efficiency of SAX make it suitable for resource-constrained devices.

Ramachandra \textit{et al.} \cite{raghavendra2014BSIF_palm} employed a series of BSIF filters that were trained for texture description on a large dataset of images. The ROI is convolved with the bank of filters and then binarized (using a specific threshold value), allowing for an 8-bit encoding.

Jia \textit{et al.} \cite{jia2013HOL} investigated the potential use of HOG \cite{dalal2005HOG}, which were successfully used in the past for robust object detection, especially pedestrians and faces. Furthermore, the Local Directional Pattern (LDP) \cite{jabid2010LDP} was evaluated in the context of palmprint feature extraction.

Zheng \textit{et al.} \cite{zheng2016don} described the 2D palmprint ROI with a descriptor recovering 3D information, a feature entitled Difference of Vertex Normal Vectors(DoN). The DoN represents the filter response of the palmprint ROI to a specific filter containing several sub-regions (of 1 or -1) intersecting in the center of the filter (borders are made up of 0s), with various orientations. In order to match two DoN templates, a weighted sum of AND, OR and XOR operators was used.

Li \textit{et al.} \cite{li2017palmprint} extracted the Local Tetra Pattern (LTrP) \cite{murala2012local_tetra_patterns} from a palmprint image that was initially filtered with a Gabor \cite{kong2004compcode} or MFRAT \cite{jia2008rloc} filter. Only the real component from the Gabor convolution was taken into consideration, after the winner-take-all rule of $arg_{min}$ was applied at pixel level between all filter orientations. Then, block-wise histograms of the LTrP values were concatenated in order to determine the final vector describing a palmprint image.\\
Wang \textit{et al.} \cite{wang2006palmprint} used the Local Binary Pattern (LBP), which encodes the value of a pixel based on a neighborhood around it \cite{ojala2002lbp}. % (by comparing the central pixel with its neighbors, the value 1 or 0 being assigned if the neighboring pixel is greater or lower than the central pixel)
Generally, the 3x3 kernel is used, allowing codes that range in value from 0 to 255.

An overview of these approaches is detailed in Table \ref{feat_overview_conventional} under category (A0).

%\textbf{*** Introduce overview (Table) of \underline{A1} approaches.***\\}
\begin{table*}
	\caption{Overview of (A1) approaches encoding the orientation at pixel level, (A2) approaches encoding the orientation at region level, and (B) approaches based on rotation/scale invariant image descriptors.} \label{feat_overview_conventional}
	\centering
	\scriptsize %\footnotesize
	%\resizebox{\columnwidth}{!}{%
	\begin{tabular}{|p{5pt}|p{15pt}|p{57.5pt}| p{190pt} |p{60pt}|p{60pt}|p{42.5pt}|}
		\hline \hline
		& \textbf{Year} & \textbf{Acronym} & \textbf{Short description} & \textbf{Classifier} & \textbf{ DB(s)} & \makecell{\textbf{Best Result}\\ \textbf{(EER/RR)}} \\
		\hline \hline
		
		\multirow{7}{5pt}{\textbf{A0}}& 2010 & SAX \cite{chen2010SAX} & Discretization of a 2D grayscale image & & \makecell[l]{HKPU\cite{TheHongKongPolytechnicUniversity}\\CASIA\cite{CASIA}} & \makecell[l]{0.3\%\\0.9\%}\\
		\cline{2-7}%\hline
		
		& 2014 & BSIF \cite{raghavendra2014BSIF_palm} & Encoding filter responses from several BSIF filters & Sparse Repr. Classifier & \makecell[tl]{HKPU\cite{TheHongKongPolytechnicUniversity}\\IITD (L, R)\cite{IITDelhi2014}} & \makecell[tl]{6.19\%\\0.42\%; 1.31\%}\\
		\cline{2-7}%\hline
		& 2014 & HOG \cite{jia2013HOL} & Histogram of Oriented Gradients & L2 distance & HKPU\cite{TheHongKongPolytechnicUniversity} & 98.03\%\\
		\cline{2-7}%\hline
		& 2016 & LDP \cite{luo2016LLDP} & Convolution with Kirsch edge masks & Manhattan + Chi-square & \makecell[tl]{HKPU\cite{TheHongKongPolytechnicUniversity}\\IITD\cite{IITDelhi2014}} & \makecell[tl]{6.10\%\\10.42\%}\\
		\cline{2-7}%\hline
		
		& 2016 & DoN \cite{zheng2016don} & 3D recovered descriptor from 2D image & weighted sum of 3 scores & \makecell[l]{HKPU\cite{TheHongKongPolytechnicUniversity}\\IITD\cite{IITDelhi2014}\\CASIA\cite{CASIA}} & \makecell[l]{0.033\%\\0.68\%\\0.53\%}\\
		\cline{2-7}%\hline  
		
		& 2017 & LBP \cite{li2017palmprint} & Local Binary Pattern &   & \makecell[l]{HKPU\cite{TheHongKongPolytechnicUniversity}\\IITD\cite{IITDelhi2014}} & \makecell[l]{4.92\%\\9.71\%}\\
		\cline{2-7}%\hline
		
		& 2017 & LTrP \cite{li2017palmprint} & Local Tetra Pattern &  & \makecell[l]{BERC1, BERC2\cite{Kim2015}\\IITD\cite{IITDelhi2014}} & \makecell[l]{1.49\%; 1.83\%\\0.94\%}\\
		\hline \hline
		
		\multirow{12}{12.5pt}[-3em]{\textbf{A1}}& 2003 & PalmCode \cite{zhang2003online} & Real and imaginary components of convolution with Gabor filter $ \pi/4 $ & Normalized Hamming & HKPU-v1 \cite{TheHongKongPolytechnicUniversity} & 0.60\% \\
		\cline{2-7}%\hline
		& 2004 & CompCode \cite{kong2004compcode} & Real components of convolution with Gabor filters (6 orientations) & Angular Distance& HKPU-v1\cite{TheHongKongPolytechnicUniversity} & 98\% at $10^{-6}$ FAR\\
		\cline{2-7}%\hline
		& 2005 & OLOF \cite{sun2005olof} & Convolution with difference of orthogonal Gaussians & Hamming & HKPU-v1\cite{TheHongKongPolytechnicUniversity} & 0.0\%\\
		\cline{2-7}%\hline
		
		& 2006 & DoG \cite{wu2006dog_code} & Convolution with Derivative of Gaussians & Hamming & HKPU \cite{TheHongKongPolytechnicUniversity} & 0.19\%\\
		\cline{2-7}%\hline
		& 2008 & RLOC \cite{jia2008rloc} & Convolution with 6 MFRAT filters & & HKPU\cite{TheHongKongPolytechnicUniversity} & 0.40\%\\
		\cline{2-7}%\hline
		& 2009 & BOCV \cite{guo2009BOCV} & Thresholding Gabor filter response. Binary encoding & Hamming &HKPU\cite{TheHongKongPolytechnicUniversity}& 0.019\% \\
		\cline{2-7}%\hline
		
		& 2011 & Contour-Code \cite{khan2011contourcode} & Two-sequence convolution, followed by hashing & Hash Table & \makecell[l]{HKPU-MS\cite{zhang2009hkpu_ms}\\CASIA-MS\cite{hao2008casia_ms}} & \makecell[l]{0.006\%\\0.3\%}\\
		\cline{2-7}%\hline
		& 2012 & E-BOCV \cite{zhang2012EBOCV} & Removing 'fragile' bits from matching& Fragile-bit pattern & HKPU\cite{TheHongKongPolytechnicUniversity} & 0.0316\%\\
		\cline{2-7}%\hline
		
		& 2016 & DOC \cite{fei2016DOC} & Include the 2 strongest orientations at pixel-level & Non-linear Angular Distance & \makecell[tl]{HKPU\cite{TheHongKongPolytechnicUniversity}\\IITD\cite{IITDelhi2014}} & \makecell[tl]{0.0092\%\\0.0622\%}\\
		\cline{2-7}%\hline
		
		%		2016 & DoN \cite{zheng2016don} & 3D recovered descriptor from 2D image & weighted sum of 3 scores & \makecell[l]{HKPU\cite{TheHongKongPolytechnicUniversity}\\IITD\cite{IITDelhi2014}\\CASIA\cite{CASIA}} & \makecell[l]{0.033\%\\0.68\%\\0.53\%}\\
		%		\hline
		
		& 2016 & Fast-RLOC \cite{zheng2016fast_compcode} & Convolution with orthogonal pairs of Gabor/MFRAT filters & Hamming & HKPU\cite{TheHongKongPolytechnicUniversity} &0.041\% \\
		\cline{2-7}%\hline
		
		& 2016 & Half-orientation Code \cite{fei2016half} & Convolution with 2 pairs of half-Gabor filters. Using both halves during matching &  & \makecell[l]{HKPU\cite{TheHongKongPolytechnicUniversity}\\IITD\cite{IITDelhi2014}} & \makecell[l]{0.0204\%\\0.0633\%}\\
		\cline{2-7}%\hline
		
		& 2017 & COM \cite{tabejamaat2017concavity_banana} & Convolution with filters describing concavity & Angular Hamming Dist. & HKPU-v2 \cite{TheHongKongPolytechnicUniversity} & 0.14\%\\
		
		\hline \hline
		
		\multirow{9}{12.5pt}[-6em]{\textbf{A2}}& 2013 & HOL \cite{jia2013HOL} & Block-wise histogram of strongest orientation & L2 distance & \makecell[l]{HKPU\cite{TheHongKongPolytechnicUniversity}\\HKPU-MS(B)\cite{zhang2009hkpu_ms}} & \makecell[l]{0.31\%\\0.064\%}\\
		\cline{2-7}%\hline
		
		& 2015 & \cite{Kim2015} & Modified CompCode (+HOG) & Chi-square distance & \makecell[l]{BERC-DB1\cite{Kim2015}\\BERC-DB2\cite{Kim2015}\\HKPU\cite{TheHongKongPolytechnicUniversity}\\IITD\cite{IITDelhi2014}} & \makecell[l]{2.88\%\\3.15\%\\0.11\%\\5.19}\\
		\cline{2-7}%\hline
		
		& 2016 & \cite{fei2016neighb_direction_indicator} & Neighboring direction indicator & &\makecell[l]{HKPU\cite{TheHongKongPolytechnicUniversity}\\IITD\cite{IITDelhi2014}}&\makecell[l]{0.0225\%\\0.0626\%} \\
		\cline{2-7}%\hline
		
		& 2016 & LLDP \cite{luo2016LLDP} & Extended encoding strategies to Gabor/MFRAT & Manhattan, Chi-square distance & \makecell[l]{HKPU\cite{TheHongKongPolytechnicUniversity}\\IITD\cite{IITDelhi2014}} & \makecell[l]{0.021\%\\4.07\%}\\
		\cline{2-7}%\hline
		
		& 2016 & LMDP \cite{fei2016LMDP} & Block-wise encoding of multiple dominant orientations &  & \makecell[l]{HKPU-v2\cite{TheHongKongPolytechnicUniversity}\\IITD\cite{IITDelhi2014}\\GPDS\cite{ferrer2011GPDS}} & \makecell[l]{0.0059\%\\0.0264\%\\0.1847\%}\\
		\cline{2-7}%\hline
		
		& 2016 & DRCC \cite{xu2016robust_compcode} & CompCode with side orientations in weighted manner & Modified Angular Distance & \makecell[l]{HKPU\cite{TheHongKongPolytechnicUniversity}\\IITD\cite{IITDelhi2014}} & \makecell[l]{0.0189\%\\0.0626\%}\\
		\cline{2-7}%\hline
		
		& 2017 & LMTrP \cite{li2017palmprint} & Local micro-tetra pattern &         & \makecell[l]{BERC1\cite{Kim2015}\\BERC2\cite{Kim2015}\\HKPU-MS\cite{zhang2009hkpu_ms}\\IITD\cite{IITDelhi2014}} & \makecell[l]{1.11\%\\1.68\%\\0.0006\%\\0.87\%}\\
		\cline{2-7}%\hline
		
		& 2017 & CR-CompCode \cite{ZHANG2017199} & Block-wise histogram of CompCode & CRC & Tongji \cite{ZHANG2017199} & 98.78\%\\
		\cline{2-7}%\hline
		
		& 2018 & CDR \cite{jia2017complete_code} & Convolution with MFRAT at several scales (15) with 12 orientations. 6 overlapping regions & BLPOC & \makecell[l]{HKPU-v2\cite{TheHongKongPolytechnicUniversity}\\HFUT\cite{xiao2019palm_roi}} & \makecell[l]{0.001\%\\0.0868\%}\\
		
		\hline \hline
		
		\multirow{6}{12.5pt}[-2em]{\textbf{B}} & 2008 & \cite{chen2008sift_pp} & SIFT + SAX. Rank-level fusion  &  & HKPU \cite{TheHongKongPolytechnicUniversity} & 0.37\% \\
		\cline{2-7}%\hline
		
		& 2011 & \cite{Morales2011} & modified SIFT (OLOF) & Similarity + Hamming & \makecell[l]{IITD\cite{IITDelhi2014}\\GPDS\cite{ferrer2011GPDS}} & \makecell[l]{0.21\%\\0.17\%}\\
		\cline{2-7}%\hline
		
		& 2013 & \cite{Zhao2013sift_iransac} & SIFT + Iterative RANSAC & I-RANSAC & IITD\cite{IITDelhi2014} & \makecell[l]{(L) 0.513\%\\(R)0.552\%} \\ 
		\cline{2-7}%\hline
		
		& 2014 & \cite{kang2014mod_sift} & RootSIFT & hierarchical matching & CASIA-MS\cite{hao2008casia_ms} & 1.00\%\\
		\cline{2-7}%\hline
		
		& 2016 & \cite{tiwari2016orb_pp} & SIFT and ORB descriptors & disimilarity index & Tiwari \cite{tiwari2016orb_pp} & 5.55\% \\
		\cline{2-7}%\hline
		
		& 2016 & \cite{charfi2016sift_svm_fusion} & Sparse representation, fused at rank-level with SVM & SRC + SVM & \makecell[l]{REST\cite{charfi2016sift_svm_fusion}\\CASIA\cite{CASIA}} & \makecell[l]{98.33\%\\99.72}\\
		\hline \hline
	\end{tabular} %}
\end{table*}

\subsubsection{Encoding Palmprint Line Orientation at Pixel Level}

One of the first approaches to extract the palmprint features from an ROI relied on only one Gabor filter oriented at $\frac{\pi}{4}$, entitled PalmCode \cite{zhang2003online}. Three values were used in the matching stage of PalmCode, namely the real, imaginary, as well as a segmentation mask to reduce the influence of poor ROI segmentation. Several approaches following a similar rationale were proposed in the following years after PalmCode, with the introduction of Competitive Code (CompCode)  \cite{kong2004compcode} and Robust Line Orientation Code (RLOC) \cite{jia2008rloc}. Both CompCode and RLOC used a competitive rule ($arg_{min}$) between a bank of filters having 6 orientations. Every pixel from the palmprint ROI was considered to be part of a line, and as the lines in the palmprint correspond to black pixels, the minimum response was chosen. Whereas CompCode used the filter response from Gabor filters, RLOC used the filter response from a modified filter Jia \textit{et al.} called MFRAT because it was inspired from the RADON transform. In the case of CompCode only the real component was used.

Gaussian filters were also used, either the derivative of two 2D Gaussian distributions (DoG \cite{wu2006dog_code}) or as the difference between two 2D orthogonal Gaussian filters (OLOF \cite{sun2005olof}).

Guo \textit{et al.} \cite{guo2009BOCV} introduced Binary Orientation Co-occurrence Vector (BOCV), obtained the filter response of a Gabor filterbank and encoded every pixel relative to a specific threshold (0 or another threshold, chosen based on the distribution of values after convolution with a specific filter). Every filter response was L1 normalized prior to the encoding, after which the thresholded values from each orientation were used to encode an 8-bit number corresponding to every pixel. An extension of this approach was introduced by Zhang \textit{et al.} \cite{zhang2012EBOCV} with EBOCV, which included masking the 'fragile' bits obtained after convolution with the Gabor filter-bank (as performed previously on IrisCode \cite{hollingsworth2008iris_code_fragile_bits} in the context of iris recognition). In this context, a 'fragile' bit is interpreted as being the pixels close to 0 (after convolution).

Khan \textit{et al.} \cite{khan2011contourcode} introduced ContourCode, obtained by convolving the input ROI in two distinct stages. Initially, the filter response corresponding to a Non-subsampled Contourlet Transform (uniscale pyramidal filter) was obtained, after which the ROI was convolved with a directional filter bank. The strongest sub-band was determined ($arg_{max}$) and the resulting code was binarized into a hash table structure.\\
Fei \textit{et al.} \cite{fei2016DOC} introduced the Double-orientation Code (DOC) which encodes the two lowest responses (to a Gabor filter bank). In order to compute the distance between two ROIs, a non-linear angular distance, measuring the dissimilarity of the two responses was determined.

Zheng \textit{et al.} \cite{zheng2016fast_compcode} investigated the effect of number of filter orientations on the efficiency of CompCode \cite{kong2004compcode} and RLOC \cite{jia2008rloc}. A single orthogonal pair of Gabor and MFRAT filters was found to perform better than when using 6 orientations. This encoding approach was called Fast-Compcode/Fast-RLOC due to its increase in speed, mostly due to a reduction in complexity. %These pairs of encoding ($arg_{min}$) the filter responses corresponding to two orthogonal filters

An interesting approach was introduced by Tabejamaat \textit{et al.} \cite{tabejamaat2017concavity_banana}, who described the concavity of a 2D palmprint ROI by convolving it with several Banana wavelet filters \cite{peters1997banana_wavelet}. Three pairs of filters (positive and negative concavity) were convolved with the ROI and a competitive rule ($arg_{min}$) was used for encoding. The joint representation was called Concavity Orientation Map (COM). An angular hamming distance was then used for matching COMs.

An overview of these approaches is detailed in Table \ref{feat_overview_conventional} under category (A1).

\subsubsection{Region-based Palmprint Line Orientation Encoding}

Jia \textit{et al.} \cite{jia2013HOL} introduced an analysis of region-based methods applied to palmprint recognition. They extended the RLOC encoding capabilities to the region-level by using the histogram of dominant orientations (after the $arg_{min}$ rule). The histograms of orientations were then concatenated. This approach essentially replaced the gradient information used in HOG with the dominant MFRAT filter response. For matching two palmprint templates, the L2 distance was used.\\
Zhang \textit{et al.} \cite{ZHANG2017199} used a similar approach to retrieve the block-wise histograms of CompCode orientations, but a Collaborative Representation Classifier (CRC) was used to perform the classification.

Kim \textit{et al.} \cite{Kim2015} used a modified version of CompCode, where a segmentation map was first determined by using the real values of the filter responses. This segmentation map was then used to compute the strongest gradients and compute the corresponding HOG. The Chi-square distance was used for matching palmprint templates.

Li \textit{et al.} \cite{li2017palmprint} extended the general approach of Local Tetra Patterns \cite{murala2012local_tetra_patterns} by replacing the derivative along the width and length with  the filter response to MFRAT \cite{jia2008rloc} or Gabor \cite{kong2004compcode} filter banks. Furthermore, the encoding method was modified to take into account the thickness of the palm lines. The image was then separated into regions and histograms were computed for each region. Finally, they were concatenated and passed through a Kernel PCA filter to reduce the dimensionality of the template.

Luo \textit{et al.} \cite{luo2016LLDP} introduced the Local Line Directional Pattern (LLDP), which represented an extension of general region encoding approaches (LDP \cite{jabid2010LDP}, ELDP \cite{zhong2013ELDP} and LDN \cite{rivera2012LDN}). The convolution stage replaced the use of Kirsch filters with Gabor or MFRAT filter banks. This step corresponds to replacing the general gradient information in a region with palmprint-specific line information. A similar approach was employed by Fei \textit{et al.} \cite{fei2018complete_binary_repr} to encode the 2D information in the context of a 3D palmprint recognition system. The response to the Gabor bank of filters was encoded using the LBP \cite{ojala2002lbp} strategy. The system used a feature-level fusion technique.\\
Fei \textit{et al.} \cite{fei2016LMDP} introduced the Local Multiple Directional Pattern (LMDP) as a way of representing two strong line orientations when these were present, instead of choosing only the dominant line orientation. The block-wise histograms of LMDP codes were computed and matching was performed using the Chi-square distance. In a similar manner, Xu \textit{et al.} \cite{xu2016robust_compcode} introduced SideCode as a robust form of CompCode, representing a combination of the dominant orientation with the side orientations in a weighted manner.\\
Fei \textit{et al.} \cite{fei2016neighb_direction_indicator} used the Neighboring Direction Indicator (NDI) to determine the dominant orientation for each pixel, along with its relation to the orientations of the neighboring regions in the image.

Jia \textit{et al.} \cite{jia2017complete_code} introduced the Complete Directional Representation (CDR) code, encoding the line orientation information at 15 scales with 12 MFRAT filters. From these images 6 overlapping regions were extracted, resulting in 1080 regions. These features were then matched using Band Limited Phase-only Correlation (BLPOC) \cite{iitsuka2008BLPOC_palm}. This approach was based on the average cross-phase spectrum of the 2D Fast Fourier Transforms (FFT) corresponding to two palmprint templates. The impulse centered on $(x_0, y_0)$ corresponds to the probability of the two templates belonging to the same class (large if intra-class, low if inter-class).

An overview of these approaches is detailed in Table \ref{feat_overview_conventional} under category (A2).

\subsubsection{Image Descriptors used for Palmprint Feature Extraction} \label{pp_feat_image_descriptors}

%\textbf{*** Introduce overview (Table) of \underline{B} approaches.***}
%\begin{table}
%	\caption{Overview of approaches based on rotation/scale invariant image descriptors.} \label{feat_overview_standard_b}
%	\centering
%	\footnotesize
%	\resizebox{\columnwidth}{!}{%
%	\begin{tabular}{|p{15pt}|p{35pt}| p{125pt} |p{60pt}|p{60pt}|p{55pt}|}
%		\hline \hline
%		\textbf{Year} & \textbf{Acronym} & \textbf{short descr.} & \textbf{classifier} & \textbf{ DB(s)} & \makecell{\textbf{Best Result}\\ \textbf{(RR/EER)}} \\
%		\hline \hline
%		
%		2008 & \cite{chen2008sift_pp} & SIFT + SAX. Rank-level fusion  &  & HKPU \cite{TheHongKongPolytechnicUniversity} & 0.37\% \\
%		\hline
%		
%		2011 & \cite{Morales2011} & modified SIFT (OLOF) & Similarity + Hamming & \makecell[l]{IITD\cite{IITDelhi2014}\\GPDS\cite{ferrer2011GPDS}} & \makecell[l]{0.21\%\\0.17\%}\\
%		\hline
%		
%		2013 & \cite{Zhao2013sift_iransac} & SIFT + Iterative RANSAC & I-RANSAC & IITD\cite{IITDelhi2014} & \makecell[l]{(L) 0.513\%\\(R)0.552\%} \\ 
%		\hline
%		
%		2014 & \cite{kang2014mod_sift} & RootSIFT & hierarchical matching & \textbf{CASIA-MS} & 1.00\%\\
%		\hline
%		
%		2016 & \cite{tiwari2016orb_pp} & SIFT and ORB descriptors & disimilarity index & Tiwari \cite{tiwari2016orb_pp} & 5.55\% \\
%		\hline
%		
%		2016 & \cite{charfi2016sift_svm_fusion} & Sparse representation, fused at rank-level with SVM & SRC + SVM & \makecell[l]{\textbf{REST}\\CASIA\cite{CASIA}} & \makecell[l]{98.33\%\\99.72}\\
%		\hline \hline
%	\end{tabular}}
%\end{table}

Image descriptors such as the Scale Invariant Feature Transform (SIFT) \cite{Lowe2004SIFT} represented a major breakthrough for object detection in unconstrained conditions because of the rotation and scale invariance of SIFT key-points.
This brought much interest to SIFT descriptors, which were either applied directly to palmprint images, such as in \cite{Zhao2013sift_iransac}, \cite{tiwari2016orb_pp}, \cite{charfi2014bosph_sift} or with certain modifications brought to one of its stages.\\
Morales \textit{et al.} \cite{Morales2011} replaced the DoG with the Ordinal Line Oriented Feature (OLOF) in the stage associated to key-point detection. Furthemore, the score determined from matching SIFT descriptors was fused with the OLOF matching prediction, making the prediction more robust. Zhao \textit{et al.} \cite{Zhao2013sift_iransac} improved the initial key-point detection stage by filtering the palmprint image with a circular Gabor filter. Then the corresponding SIFT descriptors were matched using a modified version of the RANSAC algorithm which used several iterations.

Kang \textit{et al.} \cite{kang2014mod_sift} introduced a modified SIFT which is more stable, called RootSIFT. Furthermore, histogram equalization of the graylevel image was added as a pre-processing stage. A mismatching removal algorithm (of SIFT descriptors) based on neighborhood search and LBP histograms further reduced the number of out-liers. 

Charfi \textit{et al.} \cite{charfi2016sift_svm_fusion} used a sparse representation of the SIFT descriptors to perform the matching, as well as rank-level fusion with an SVM. Similarly, a rank-level fusion was performed by Chen \textit{et al.} \cite{chen2008sift_pp} matching SAX and SIFT descriptors. 

Tiwari \textit{et al.} matched SIFT and ORB \cite{rublee2011ORB} descriptors acquired using smartphone cameras. As with most other approaches using SIFT descriptors, a dissimilarity function was defined, counting the number of in-lier matches performed between two images. Srinivas \textit{et al.} \cite{srinivas2009palmprint_surf} used Speeded Up Robust Features (SURF) \cite{Bay2008} to match two palmprint ROIs. They further improved the matching speed by only matching the SURF descriptors extracted from specific subregions of the ROI, instead of the entire surface of the ROI.

An overview of these approaches is detailed in Table \ref{feat_overview_conventional} under category (B).
\subsection{CNN-based Approaches} \label{section_cnn_based_feature_extract}

One of the great advantages of using CNNs is that the filters are learned from a specific training distribution, which makes them relevant to the task of palmprint recognition. As opposed to traditional (crafted) features, the learned features are trained to describe any distribution. The main disadvantage of this approach lies in the requirement of abundant and accurately labeled training data, which generally is a problem.

The existing approaches for palmprint feature extraction relying on CNNs, can be split into three categories:
\begin{itemize}
	\item Using pre-trained models (on ImageNet), the network's output is considered to be the extracted feature. Also relies on a classifier such as SVM.
	\item Networks of filters optimised using various approaches.
	\item Training from scratch (or using transfer-learning) of DNNs to determine embeddings that minimize intra-class distance and maximize inter-class distance.
\end{itemize}

\begin{table*}
	\caption{Pre-trained networks (C1), or linear Neural Networks (C2). Training CNNs for palmprint feature extraction (C3A). Siamese approaches (C3B) to training CNNs for palmprint feature extraction.} \label{feat_overview_nn}
	\centering
	\scriptsize %\footnotesize
	%\resizebox{\columnwidth}{!}{%
	\begin{tabular}{|p{12.5pt}|p{15pt}|p{52.5pt}| p{170pt} |p{60pt}|p{60pt}|p{42.5pt}|}
		\hline \hline
		&\textbf{Year}& \textbf{Acronym} & \textbf{Short description} & \textbf{Classifier} & \textbf{ DB(s)} & \makecell{\textbf{Best Result}\\ \textbf{(EER/RR)}} \\
		\hline \hline
		
		\multirow{3}{20pt}[0em]{\textbf{C1}} & 2016 & \cite{dian2016palmprint_cnn} & Log-its layer of AlexNet, pre-trained on ImageNet  & Hausdorff dist. & \makecell[l]{HKPU\cite{TheHongKongPolytechnicUniversity}\\ CASIA\cite{CASIA}\\ IITD\cite{IITDelhi2014}} & \makecell[l]{0.044\%\\0.0803\%\\0.1113\%} \\
		\cline{2-7}%\hline
		
		& 2018 &  \cite{tarawneh2018} & Output of FC6 layer from AlexNet, VGG16/19, pre-trained on ImageNet & SVM & \makecell[l]{MOHI \cite{hassanat2015mohi}\\ COEP\cite{coep_palmprint_dataset}} & \makecell[l]{95.50\% \\ 98.00\%}\\
		\cline{2-7}%\hline
		
		& 2018 &  \cite{ramachandra2018icb} & Transfer learning: AlexNet pre-trained on ImageNet & fusion: Softmax+SVM) & CPNB \cite{ramachandra2018icb} & 0.310\% \\
		
		\hline \hline
		
		\multirow{3}{20pt}[-3em]{\textbf{C2}} & 2016 & \textbf{ScatNet} \cite{minaee2016palmprint} & Deep Scattering Network, fixed weights & \makecell[l]{lin-SVM\\K-NN} & HKPU \cite{TheHongKongPolytechnicUniversity} &  \makecell[l]{100\% \\ 94.40\%} \\
		\cline{2-7}%\hline
		
		& 2016 & \textbf{PCANet} \cite{meraoumia2016palm_pcanet} & Obtaining filters based on PCA and training distribution & SVM & \makecell[l]{HKPU-MS \cite{zhang2009hkpu_ms}\\ CASIA-MS \cite{hao2008casia_ms}} & \makecell[l]{0.0\% \\ 0.12\%} \\
		\cline{2-7}%\hline
		
		&2019 & \textbf{PalmNet} \cite{genovese2019palmnet}  & Modification of PCANet, with the 2nd layer composed of Gabor filters (selected adaptively, based on a training distribution)  & 1-NN, L2 dist. &  \makecell[tl]{CASIA\cite{CASIA} \\ IITD\cite{IITDelhi2014} \\ REST\cite{charfi2016sift_svm_fusion} \\ Tongji \cite{zhang2015collaborative_3d_pp}} & \makecell[tl]{0.72\% \\ 0.52\% \\ 4.50\% \\ 0.16\%}\\
		\hline \hline
		
		%\hline \hline
		%\textbf{Year} &\textbf{Cat.}& \textbf{Acronym} & \textbf{Short description} & \textbf{classifier} & \textbf{ DB(s)} & \makecell{\textbf{Best Result}\\ \textbf{(RR/EER)}} \\ 
		%\hline 
		
		\multirow{8}{20pt}[-4em]{\textbf{C3-A}} & 2015 & \cite{jalali2015cnn} & Shallow net & Softmax & \makecell[l]{HKPU-MS \cite{zhang2009hkpu_ms}\\Own} & \makecell[l]{99.97\%\\93.4\%}\\
		\cline{2-7}%\hline
		
		%& 2015 & \cite{xin2015palmprint} & Shallow net & Softmax & Own &  90.60\% \\
		%\cline{2-7}%\hline
		
		& 2017 & \cite{afifi201711k} & Two-stream CNN: low-frequency and high-frequency, then trained to classify the image according to its class. & SVM & \makecell[l]{11KHands\cite{afifi201711k}\\ IITD\cite{IITDelhi2014}} & \makecell[l]{96.00\%\\94.80\%} 
		\\
		\cline{2-7}%\hline
		
		& 2018 & \cite{Izadpanahkakhk2018} & Transfer-learning: CNN \cite{chatfield2014return} pre-trained on ImageNet, re-trained with cross-entropy & KNN, SVM, RFC & \makecell[l]{HKPU\cite{TheHongKongPolytechnicUniversity}\\ IITD\cite{IITDelhi2014}} & \makecell[l]{100\%\\96.9\%}\\
		\cline{2-7}%\hline
		
		& 2018 & \textbf{Palm-RCNN}\cite{zhang2018palmprint} & Inception-ResNetV1, with Cross-entropy and Center loss (combined loss) & SVM, L2 dist. & Tongji-MS\cite{zhang2018palmprint} & \makecell[l]{100\%*\\2.74\%***}\\
		\cline{2-7}%\hline\hline
		
		& 2019 & \cite{fei2018pp_feat_extr_review} & Transfer-learning: AlexNet, VGG16, InceptionV3 and ResNet50, retrained with cross-entropy loss & Softmax &\makecell[tl]{CASIA\cite{CASIA}\\IITD\cite{IITDelhi2014}\\GPDS-CL1\cite{ferrer2011GPDS}} &\makecell[tl]{3.78\% \\4.79\% \\4.69\%}\\
		\cline{2-7}%\hline		
		
		& 2019 & \textbf{JDCFR} \cite{zhao2019joint_cnn_palm} & Shallow CNNs trained on each spectral band (53) & CRC & Own (MS) & 0.01\% \\
		\cline{2-7}%\hline
		
		& 2019 & \cite{izadpana2019novel_mobile_datasets} & Transfer learning: VGG16, GoogLeNet, \cite{chatfield2014return} pre-trained on ImageNet; trained using Cross-entropy loss & Softmax &  SMPD\cite{izadpana2019novel_mobile_datasets} & 93.40\%\\
		
		\cline{2-7}%\hline
		& 2019 & \makecell[tl]{\textbf{FERnet},\\ \textbf{EE-PRnet}\cite{matkowski2019palmprint}} &Architecture based on pre-trained VGG16 (pruned after 3rd maxpool),'' with \textbf{D} and \textbf{FC} (FERnet). EE-PRnet is trained end-to-end, together with the ROI extraction architecture. Trained with cross-entropy loss. & PLS\cite{geladi1986PLS}&\makecell[tl]{NTU-PI\cite{matkowski2019palmprint}\\NTU-CP\cite{matkowski2019palmprint}; IITD\cite{IITDelhi2014}\\HKPU\cite{TheHongKongPolytechnicUniversity}; CASIA\cite{CASIA}} & \makecell[tl]{41.92\%;64.73*\\ 0.76\%;  0.26\%\\0.15\%; 0.73\%}\\
		%
		
		%\hline \hline
		%2016 & C3-B &\cite{svoboda2016palmprint} & Siamese network trained with d-prime loss & distance & \makecell[l]{CASIA\cite{CASIA}\\IITD\cite{IITDelhi2014}} & \makecell[l]{1.86\%\\1.64\%} \\
		
		\hline \hline
		
		\multirow{8}{20pt}[-7em]{\textbf{C3-B}} &2016 & \cite{svoboda2016palmprint} & Siamese network trained with d-prime loss &  & \makecell[l]{CASIA\cite{CASIA}\\IITD\cite{IITDelhi2014}} & \makecell[l]{1.86\%\\1.64\%} \\
		\cline{2-7}%\hline\hline
		
		&2018 & \textbf{RFN} \cite{liu2020shifted_triplet_loss} & Soft-shifted Triplet loss &  & \makecell[l]{IITD\cite{IITDelhi2014}\\ PolyU-IITD\cite{kumar2018HKPU_IITD_v3} } & \makecell[l]{0.68\%\\0.15\%}\\
		\cline{2-7}%\hline\hline
		
		& 2018 & \cite{zhong2018siamese} & VGG16 retrained last layers &  & \makecell[l]{HKPU\cite{TheHongKongPolytechnicUniversity}\\ XJTU-UP\cite{shao2019efficient}} & \makecell[l]{0.2819\%\\4.559\%}\\
		\cline{2-7}%\hline\hline
		
		& 2019 & \textbf{DHCN} \cite{shao2019efficient} & Binary Hashing, with Knowledge Distillation \cite{hinton2015distilling} &  & \makecell[l]{XJTU-UP\cite{shao2019efficient}\\ XJTU-kd \cite{shao2019efficient}} & \makecell[l]{0.60\%\\5.83\%}\\
		\cline{2-7}%\hline\hline
		
		& 2019 & \textbf{Deep-MV} \cite{zhang2019palm_roi}& MobileNetV2 with secondary net & - & MPD \cite{zhang2019palm_roi} & 89.91\% \\
		\cline{2-7}%\hline\hline
		
		& 2019 & \textbf{PalmGAN} \cite{shao2019palmgan} & Cross-domain transformation &  & \makecell[l]{HKPU-MS \cite{zhang2009hkpu_ms}\\ SemiU\cite{shao2019palmgan}\\Uncontr. \cite{shao2019palmgan}} &\makecell[l]{0.0\%\\0.005\%\\5.55\%}\\ 
		\cline{2-7}%\hline\hline
		
		& 2019 & \cite{du2019low_shot} & Siamese with secondary network. Few-shot training & Softmax & \makecell[l]{HKPU-MS \cite{zhang2009hkpu_ms}\\Pa, Pb\\Pc, Pd}& \makecell[l]{99.4\%\\95.4\%, 93.4\%\\98.8\%, 96.4\%}\\
		\hline	
		
	\end{tabular}\\
	\textbf{D} = dropout; \textbf{FC} = fully connected layers; * and ** refer to the identification results expressed in Recognition Rate, Rank-1 and Rank-30. *** refers to verification results, as opposed to identification.%evaluation of EE-PRnet\cite{matkowski2019palmprint} on NTU-PI-v1 dataset, which correspond to Rank-1 and Rank-30 respectively.%}
\end{table*}

\subsubsection{Using pre-trained DNNs}
Dian \textit{et al.} \cite{dian2016palmprint_cnn} used AlexNet \cite{krizhevsky2012alexnet} pre-trained on ImageNet to extract deep features. These were then matched using the Hausdorff distance. 
In a similar fashion, Tarawneh \textit{et al.} \cite{tarawneh2018} used several networks pretrained on ImageNet (AlexNet, VGG16 \cite{VGG16_initial} and VGG19). The extracted deep features from the images in two hand datasets (COEP \cite{coep_palmprint_dataset} and MOHI \cite{hassanat2015mohi}) were then matched using a multi-class SVM.\\
Ramachandra \textit{et al.} \cite{ramachandra2018icb} used transfer-learning (AlexNet) to match palmprints acquired from infants. The class decision was obtained through a fusion rule, which took into consideration the prediction from an SVM, as well as the Softmax prediction of the network. %A novel dataset of palmprints (CPNB) was developed for these experiments, but 

An overview of these approaches is presented in Table \ref{feat_overview_nn} under category (C1).

\subsubsection{PCANet, ScatNet and PalmNet}
%Scattered network and PCANet, PalmNet(Genovese)
Minaee \textit{et al.} \cite{minaee2016palmprint} employed a scattering network (ScatNet) that was first introduced by Bruna \textit{et al.} \cite{bruna2013scattering_networks} for pattern recognition tasks, especially because of its invariance to transformations such as translation and rotation. % At each layer the local descriptor of the input signal are computed with a cascade of three operations: wavelet decompositions, complex modulus and local averaging. 
ScatNet uses Discrete Wavelet Transforms (DWT) as filters and considers the output(s) at each layer as the network outputs (not just the last layer), providing information regarding the interference of frequencies in a given image \cite{bruna2013scattering_networks}. Meraoumia \textit{et al.} used a filter bank of 5 scales and 6 orientations, the network having an architecture composed of 2 layers. The palmprint ROIs were split into blocks of 32x32 pixels and passed through the network, resulting in 12,512 scattering features. PCA was applied to reduce the dimensionality, reducing it to the first 200 components. A linear SVM was then used for the classification task.

Chan \textit{et al.} \cite{chan2015pcanet} initially introduced PCANet for general pattern recognition applications. Unlike DNNs which make use of the Rectified Linear Unit (ReLU), the PCANet does not contain any non-linearity. Instead, the filters are determined from a distribution of training images. Specifically, a series of overlapping blocks are extracted from every input image, after which the mean is removed. Based on the derived covariance matrix a number of Eigen vectors are extracted (after being sorted, the top 8) and considered as filters belonging to the first layer. The input to the second layer is the distribution of input images to the 1st layer, but convolved with the computed filters in layer 1. This process is repeated for any given number of layers, but generally architectures with 2 layers are commonplace.\\
PCANet was used for palmprint feature extraction by Meraoumia \textit{et al.} \cite{meraoumia2016palm_pcanet} on two datasets - CASIA Multispectral \cite{hao2008casia_ms} and HKPU-MS \cite{zhang2009hkpu_ms}. For classification, both SVM and KNN reported 0\% EER across all spectral bands for HKPU-MS and 0.12\% EER for CASIA-MS. However, after applying a score-fusion scheme where the first 3 bands are used, the EER drops to 0\%.

Recently, Genovese \textit{et al.} \cite{genovese2019palmnet} expanded the PCANet approach to include convolutions with fixed-size and variable-sized Gabor filters in the 2nd layer. The described architecture entitled 'PalmNet' determines the Gabor filters with the strongest response, followed by a binarization layer. An alternative architecture is considered, entitled 'PalmNet-GaborPCA', where the filters of the first layer are configured using the PCA-based tuning procedure used in PCANet, whereas the kernels in the 2nd layer are configured using the Gabor-based tuning procedure. For classification, a simple KNN classifier is used. \\
PalmNet represents an interesting approach for quickly training on large datasets of palmprints, at the same time requiring fewer resources than DNNs.

An overview of these approaches is presented in Table \ref{feat_overview_nn} under category (C2).

\subsubsection{Training DNNs}

The main distinction separating approaches in this category is the training strategy being used.\\
If the classification task is borrowed from the standard pattern recognition problem (like the ImageNet challenge), then the CNN is required to predict the class to which an input palm print belongs to. The network's last layer is fully connected with a number of units corresponding to the number of classes (in the form of a one-hot vector, depending on the size of the dataset), with the activation function being Softmax (expressing the probability of that input image to belong to either class). In this case, the loss function is the cross-entropy. Example implementations include \cite{izadpana2019novel_mobile_datasets, afifi201711k, Izadpanahkakhk2018,jalali2015cnn, fei2018pp_feat_extr_review, zhao2019joint_cnn_palm}. \\
Fei \textit{et al.} \cite{fei2018pp_feat_extr_review} compared the performance of several networks like AlexNet, VGG16, InceptionV3 and ResNet50. Izadpanahkakhk \textit{et al.} \cite{izadpana2019novel_mobile_datasets} trained and evaluated four networks (GoogLeNet, VGG16, VGG19 and a CNN developed by Chatfield \textit{et al.} \cite{chatfield2014return} for the ImageNet challenge) on two novel palmprint datasets. \\
Alternatively, after training with cross-entropy loss, the output from the log-its layer (the layer preceding the Softmax layer) can be considered as the extracted feature, which is then used to train a classifier such as SVM \cite{afifi201711k}, CRC \cite{zhao2019joint_cnn_palm} or Random Forest Classifier (RFC) \cite{Izadpanahkakhk2018}. Zhang \textit{et al.} \cite{zhang2018palmprint} used a combination of cross-entropy and center-loss functions during training for multi-spectral palmprint matching. After learning a representation of palmprints, they then fed the embeddings (output of log-its layer) to an SVM.\\
Afifi \textit{et al.} also take into consideration separating the input image's information into either high-frequency and low-frequency, thus having a two-stream CNN. The two branches later concatenate, to allow the training based on classification. Several of these layers' outputs are then concatenated, and then classified using an SVM which employs a SUM rule for fusion.

Matkowski \textit{et al.} \cite{matkowski2019palmprint} provided the first CNN-based solution for palmprint recognition which was trained End-to-End (EE-PRnet) for palmprint feature extraction. This architecture was composed of the previously mentioned ROI-LAnet and FERnet, which was also based on a pre-trained VGG16 (pruned after the 3rd Maxpool) architecture. This was followed by two fully connected  (FC) layers benefiting from Droput regularization. The network is trained using Cross-entropy (a 3rd FC layer was added to the network, corresponding to palmprint classes), but the authors explore several training scenarios regarding the Dropout layers, or fine-tune specific blocks in FERnet. Furthermore, a color augmentation protocol consisting of randomly shifting the saturation and contrast of images , was performed on-the-fly during training. \\%, which contains a localization module (trained for 2D key-point prediction using L2 loss), and a 
After obtaining the palmprint embeddings (from the 2nd FC layer), they are matched using Partial Least Squares regression (PLS) \cite{geladi1986PLS}, linear SVM, KNN-1 and Softmax. The best results were obtained using PLS.\\
Overall, the EE-PRnet provides the best results, showing that training both networks (ROI-LAnet and FERnet) together allows the architecture to reach a better understanding of the features contained in the palmprint, as well as the distortions brought by the hand's pose. Furthermore, this setup provides a considerable advantage, as the input to the network is the full image, not a cropped image of the hand.

An overview of these approaches is presented in Table \ref{feat_overview_nn} under category (C3-A).\\

Another training approach is to use the Siamese architecture (overview presented in Table \ref{feat_overview_nn}), characterized by two inputs (or several) resulting in two embeddings (usually 128 units corresponding to the last fully-connected layer) that are then compared with a loss function to determine how similar they are versus how similar they should be. This architecture, where the same network outputs the two embeddings, relies on a similarity estimation function, such as the Contrastive loss \cite{hadsell2006contrastive_loss}, or the Center loss \cite{wen2016discriminative}, where the distance between inputs is minimized (intra-class) or increased (inter-class). When the three inputs (triplets) are considered, the distance between the anchor and the positive sample is reduced while increasing the distance between the anchor and the negative sample \cite{schroff2015facenet}. \\
Svoboda \textit{et al.} \cite{svoboda2016palmprint} introduced a loss function called 'discriminative index', aimed at separating genuine-impostor distributions. Zhong \textit{et al.} \cite{zhong2018siamese} used transfer-learning based on VGG16 (initially trained on ImageNet) and Contrastive loss. %\\

Zhang \textit{et al.} \cite{zhang2019palm_roi} used a Siamese architecture of two MobileNets \cite{howard2017mobilenets} outputting feature vectors that are then fed to a sub-network tasked with the intra-class probability (0 for inter-class and 1 for intra-class, with 0.5 as a decision threshold). It is not clear, however, what loss function they used (most likely contrastive loss). Du \textit{et al.} \cite{du2019low_shot} used a similar architecture trained using the few-shot strategy. Shao \textit{et al.} \cite{shao2019neural_graphs} used the output of a 3-layer Siamese network, and matched the palmprints from two datasets (HKPU-Multispectral and a dataset collected with a smartphone camera) with a Graph Neural Network (GNN). Unfortunately, the training details of the Siamese network are not clear. \\
Liu \textit{et al.} \cite{liu2020shifted_triplet_loss} introduced the soft-shifted triplet loss as a 2D embedding specifically developed for palmprint recognition (instead of a 1D embedding). Furthermore, translations on x and y axes were used to determine the best candidates for triplet pairs (at batch level).
Recently, Shao \textit{et al.} \cite{shao2019efficient} introduced an approach based on hashing coding, where the embeddings used to encode the palmprint classes are either 0 or 1. Furthermore, similar matching performances were obtained using a much smaller network, obtained via Knowledge Distillation \cite{hinton2015distilling}. These are worthwhile directions for development, as they represent solutions to the limitations of resource-constrained devices.

%The CNN is trained so that it learns to estimate the distance between a number of inputs and maximize the distance between inter-class pairs and minimize intra-class pairs. These networks are generally called Siamese, with example loss functions including the Contrastive loss \cite{hadsell2006contrastive_loss} or the Triplet loss \cite{schroff2015facenet}.

A promising strategy for cross-device palmprint matching was recently proposed by Shao \textit{et al.} \cite{shao2019palmgan} with PalmGAN, where a cycle Generative Adversarial Network (cycle GAN) \cite{zhu2017cycleGAN} was used to perform cross-domain transformation between palmprint ROIs. A proof of concept was evaluated on the HKPU-Multispectral (HKPU-MS) palmprint dataset containing palm images acquired at several wavelengths, as well a semi-unconstrained dataset acquired with several devices.

An overview of these approaches is presented in Table \ref{feat_overview_nn} under category (C3-B).

\section{Discussion and Conclusions} \label{lit_rev_concl}
\subsection{Palmprint Datasets} \label{lit_rev_dataset_conclusions}

%\subsection{Discussion and conclusions}
% introduction of datasets, as the approaches became more and more robust
% list datasets (from first category to 2nd)
The advancement of palmprint recognition relies on the release of relevant datasets which reflect specific sets of requirements. Initially the main focus was placed on recognition, allowing little to no flexibility in terms of interaction with the system (e.g. HKPU \cite{TheHongKongPolytechnicUniversity}).\\
As the sensor technology progressed (and new consumer devices appeared on the market), there was more room for various aspects, i.e. contactless systems (IITD \cite{IITDelhi2014}, CASIA \cite{CASIA}). Then invariance to various factors of the acquisition encouraged the introduction of datasets like BERC \cite{Kim2015} (background), or 11K Hands \cite{afifi201711k} (hand pose) and PRADD \cite{Jia2012} (devices used for acquisition). Unfortunately there are several datasets that are no longer available to researchers, such as PRADD \cite{Jia2012} or DevPhone \cite{Aoyama2013}. Some recently introduced datasets are yet to be released to the research community (e.g. HFUT \cite{xiao2019palm_roi}, MPD \cite{zhang2019palm_roi} or XJTU-UP \cite{shao2019efficient}). % HFUT \cite{xiao2019palm_roi}, MPD \cite{zhang2019palm_roi}

Following the general trend of biometric recognition migrating to consumer devices, the last years have seen the introduction of several large-scale palmprint datasets (e.g. XJTU-UP \cite{shao2019efficient}) reflecting the challenging operating conditions brought by a mobile environment. \\
A new category of unconstrained palmprint datasets was recently introduced with NTU-PI-v1 \cite{matkowski2019palmprint}, including the palmprint acquired with conventional cameras to the list of forensic applications. This collection of palmprints gathered from the Internet proved to be especially challenging, given the low resolution of images, the high degree of distortion, as well as the large number of hand classes.\\
It is our opinion that these will be the most meaningful palmprint datasets for the upcoming 5-10 years, anticipating the adoption of palmprint recognition on smartphones and other devices. 
An overview of this transition was presented, the culmination of which is represented by the fully unconstrained datasets class, initiated with the introduction of NUIG\_Palm1 \cite{Ungureanu2017} in 2017.

\subsection{Palmprint ROI Extraction}
%\subsection{Discussion and conclusions}
The approaches used for palmprint region of interest extraction are linked directly with the operating conditions of devices used for acquisition. In palmprint datasets where the background is fixed (e.g. HKPU, CASIA, IITD, COEP) the task of segmentation is a straightforward procedure. However, when the background is unconstrained such as is the case with images from BERC, skin color thresholding provides limited results, even when the skin model is computed for every image based on a distribution of pixels \cite{Kim2015}. 

With the migration of palmprint recognition onto consumer devices, the general pipeline for ROI extraction needs to take into consideration more challenging factors such as lighting conditions, hand pose and camera sensor variation. It is in this context that more powerful approaches based on machine learning or deep learning can provide robust solutions without imposing strict protocols for acquisition onto the user of consumer devices. \\
%A complete evaluation of these approaches must include the following points:
A complete evaluation of these approaches is yet to be made in terms of:
%Regarding the evaluation of such region of interest extraction techniques, a number of topics that still need to be addressed are:
\begin{enumerate}
	\item The prediction error of the key-points used for ROI extraction/alignment. This seems to have been a commonly overlooked step in most research papers, with some exceptions (e.g. \cite{khan2011contourcode}).
	\item Recognition rate and the main sources of error (from the ROI extraction) affecting recognition.
	\item Running time and resource requirements, especially for CNN-based approaches. Low inference time is expected from all solutions running on consumer devices.
\end{enumerate}

%At the time of writing this literature review, there is currently one CNN-based solution for the detection and extraction of palmprint ROIs in an unconstrained environment, as introduced by Matkowski \textit{et al.} \cite{matkowski2019palmprint}. B

Furthermore, at the time of writing of this literature review, there are currently no CNN-based solutions to detect the palmprint in unconstrained environments, besides the Fast R-CNN approach demonstrated by Liu \textit{et al.} \cite{liu2020shifted_triplet_loss}, which is a Fast-RCNN.\\% which does not normalize the extracted ROI's rotation. \\
The recent use of a CNN for the normalization of palmprint ROIs regarding hand pose by Matkowski \textit{et al.} \cite{matkowski2019palmprint} has opened up exciting new possibilities for unconstrained palmprint ROI extraction (they do not address the task of palmprint detection). The Spatial Transform Network learns a non-affine transform applied to the ROI, defined by the palmprint's labeled key-points. Alternatively, pose correction could be made using 3D information, similar the work of Kanhangad \textit{et al.} \cite{kanhangad2010pose_invariant}. Although a special 3D sensor is used in \cite{kanhangad2010pose_invariant}, the hand's 3D structure can be recovered from the 2D image with hand pose estimation algorithms (as was developed by Mueller \textit{et al.} \cite{Mueller2017}). %In fact, the addition of 3D information into the Localization Network of ROI-LAnet \cite{matkowski2019palmprint} could further improve the prediction of 2D 

\subsection{Palmprint Feature Extraction}
%\subsection{Discussion and conclusions}
Although palmprint recognition took off in early 2000's with the introduction of HKPU \cite{TheHongKongPolytechnicUniversity} dataset, the pipeline stage that received the most attention from the research community has been the palmprint feature extraction. \\
% general trend: relying heavily on CNNs
As was the case for iris and face recognition, CNNs have become the current state of the art in palmprint recognition (Section \ref{section_cnn_based_feature_extract}). 
The general trend is to either train a network using Cross-entropy or Center-loss (e.g. \cite{afifi201711k, izadpana2019novel_mobile_datasets,zhang2018palmprint, fei2018pp_feat_extr_review, matkowski2019palmprint}), Siamese networks (e.g. \cite{svoboda2016palmprint, liu2020shifted_triplet_loss, zhang2019palm_roi, shao2019palmgan}), but there are  or also entirely linear networks (PCANet \cite{meraoumia2016palm_pcanet} and PalmNet \cite{genovese2019palmnet}). 
%The general trend is to use Siamese networks (e.g. variant of triplet loss in RFN \cite{liu2018shifted_triplet_loss} or contrastive loss in DHCN \cite{shao2019efficient}, DeepMPV \cite{zhang2019palm_roi}), but there are  or almost entirely linear networks (PCANet \cite{meraoumia2016palm_pcanet} and PalmNet \cite{genovese2019palmnet}). 

% they use datasets acquired with smartphones
It is important to note that most of these works use in their training/evaluation scenarios images acquired with smartphones (on datasets such as XJTU-UP \cite{shao2019efficient} and MPD \cite{zhang2019palm_roi}). The cross-device training and matching will become a main focus especially for device-independent palmprint recognition solutions, as demonstrated by \cite{matkowski2019palmprint}. This is first investigated in \cite{Ungureanu2017}, with impressive results being obtained in \cite{liu2020shifted_triplet_loss} and \cite{matkowski2019palmprint}. The cross-domain conversion of a palmprint ROI using a generative approach \cite{shao2019palmgan} also represents a promising direction of research. A GAN-based architecture might benefit from the ROI pose-normalization approach introduced by Matkowski \textit{et al.} \cite{matkowski2019palmprint}, where the ROI extraction network contains a Spatial Transform Network \cite{jaderberg2015spatialTransf}.

% complexity of architectures -> distillation and hashing is nice
The complexity of architectures becomes an important factor to optimize for devices with limited resources, as in \cite{shao2019efficient}, where the network is distilled (number of layers is reduced) and the network's output is a discrete hash code (binary values). This not only reduces the processing requirements (including matching), but also the storage space necessary when dealing with a large number of classes. An alternative approach would be to consider the ternarization of networks \cite{li2016ternary}.

% Processing time evaluation is required
As in the case of ROI extraction algorithms, the feature extraction approaches (especially the CNN-based solutions) require an evaluation in terms of processing time, as this aspect is only touched in few papers (e.g. \cite{Kim2015} and \cite{liu2020shifted_triplet_loss}).

% or


%\appendices

%\section{}

\section*{Acknowledgment}
The research work presented here is funded under Industry/Academic Partnership 13/SPP/I2868 co-funded by Science Foundation Ireland (SFI) and FotoNation Ltd.

\ifCLASSOPTIONcaptionsoff
  \newpage
\fi



%\bibliographystyle{IEEEtran}
% (used to reserve space for the reference number labels box)%

\bibliographystyle{IEEEtran}
\bibliography{references}

\begin{thebibliography}{100}
\providecommand{\url}[1]{#1}
\csname url@samestyle\endcsname
\providecommand{\newblock}{\relax}
\providecommand{\bibinfo}[2]{#2}
\providecommand{\BIBentrySTDinterwordspacing}{\spaceskip=0pt\relax}
\providecommand{\BIBentryALTinterwordstretchfactor}{4}
\providecommand{\BIBentryALTinterwordspacing}{\spaceskip=\fontdimen2\font plus
\BIBentryALTinterwordstretchfactor\fontdimen3\font minus
  \fontdimen4\font\relax}
\providecommand{\BIBforeignlanguage}[2]{{%
\expandafter\ifx\csname l@#1\endcsname\relax
\typeout{** WARNING: IEEEtran.bst: No hyphenation pattern has been}%
\typeout{** loaded for the language `#1'. Using the pattern for}%
\typeout{** the default language instead.}%
\else
\language=\csname l@#1\endcsname
\fi
#2}}
\providecommand{\BIBdecl}{\relax}
\BIBdecl

\bibitem{corcoran2015mobile_biometrics_overview}
P.~Corcoran and C.~Costache, ``Biometric technology and smartphones: A
  consideration of the practicalities of a broad adoption of biometrics and the
  likely impacts,'' in \emph{2015 IEEE International Symposium on Technology
  and Society (ISTAS)}.\hskip 1em plus 0.5em minus 0.4em\relax IEEE, 2015, pp.
  1--7.

\bibitem{amos2016openface}
B.~Amos, B.~Ludwiczuk, M.~Satyanarayanan \emph{et~al.}, ``Openface: A
  general-purpose face recognition library with mobile applications,''
  \emph{CMU School of Computer Science}, vol.~6, 2016.

\bibitem{thavalengal2015iris}
S.~Thavalengal, P.~Bigioi, and P.~Corcoran, ``Iris authentication in handheld
  devices-considerations for constraint-free acquisition,'' \emph{IEEE
  Transactions on Consumer Electronics}, vol.~61, no.~2, pp. 245--253, 2015.

\bibitem{roberts2007biom_atack_vectors}
C.~Roberts, ``Biometric attack vectors and defences,'' \emph{Computers \&
  Security}, vol.~26, no.~1, pp. 14--25, 2007.

\bibitem{jain2004introduction2Biom}
A.~K. Jain, A.~Ross, and S.~Prabhakar, ``An introduction to biometric
  recognition,'' \emph{IEEE Transactions on circuits and systems for video
  technology}, vol.~14, no.~1, pp. 4--20, 2004.

\bibitem{genovese2019civemsa}
A.~Genovese, V.~Piuri, F.~Scotti, and S.~Vishwakarma, ``Touchless palmprint and
  finger texture recognition: A deep learning fusion approach,'' in \emph{Proc.
  of the 2019 IEEE Int. Conf. on Computational Intelligence and Virtual
  Environments for Measurement Systems \& Applications (CIVEMSA 2019)},
  Tianjin, China, June 2019.

\bibitem{meraoumia2011fusion}
A.~Meraoumia, S.~Chitroub, and A.~Bouridane, ``Fusion of finger-knuckle-print
  and palmprint for an efficient multi-biometric system of person
  recognition,'' in \emph{2011 IEEE International Conference on Communications
  (ICC)}.\hskip 1em plus 0.5em minus 0.4em\relax IEEE, 2011, pp. 1--5.

\bibitem{matkowski2019study}
W.~M. Matkowski, F.~K.~S. Chan, and A.~W.~K. Kong, ``A study on wrist
  identification for forensic investigation,'' \emph{Image and Vision
  Computing}, 2019.

\bibitem{redrock_palmID}
``Epson and redrock biometrics bring first biometric authentication solution to
  consumer augmented reality headsets,'' May 2018,
  {https://www.businesswire.com/news/home/20180508005612/en/Epson-Redrock-Biometrics-Bring-Biometric-Authentication-Solution}.

\bibitem{Ungureanu2017}
\BIBentryALTinterwordspacing
A.-S. Ungureanu, S.~Thavalengal, T.~E. Cognard, C.~Costache, and P.~Corcoran,
  ``{Unconstrained palmprint as a smartphone biometric},'' \emph{IEEE
  Transactions on Consumer Electronics}, vol.~63, no.~3, pp. 334--342, aug
  2017. [Online]. Available: \url{http://ieeexplore.ieee.org/document/8103383/}
\BIBentrySTDinterwordspacing

\bibitem{TheHongKongPolytechnicUniversity}
\BIBentryALTinterwordspacing
{The Hong Kong Polytechnic University}, ``{PolyU Palmprint Palmprint Database,
  "http://www.comp.polyu.edu.hk/{\~{}}biometrics/"}.'' [Online]. Available:
  \url{http://www.comp.polyu.edu.hk/{~}biometrics/}
\BIBentrySTDinterwordspacing

\bibitem{Bosphorus_hand_database}
{Bogazici University, Istanbul, Turkey}, ``{Bosphorus Hand Database},''
  http://bosphorus.ee.boun.edu.tr/hand/Home.aspx.

\bibitem{CASIA}
{The Chinese Academy of Sciences, Automation Institute }, ``{CASIA Palmprint
  Database},'' http://biometrics.idealtest.org/.

\bibitem{IITDelhi2014}
{Indian Institute of Technology Delhi}, ``{IIT Delhi Touchless Palmprint
  Database (Version 1.0)},'' pp. 4--6, 2014,
  {URL:"http://web.iitd.ac.in/ajaykr/Database{\_}Palm.htm"}.

\bibitem{coep_palmprint_dataset}
{College of Engineering, Pune-411005(An Autonomous Institute of Government of
  Maharashtra)}, ``{Palmprint Dataset},''
  {URL:http://www.coep.org.in/resources/coeppalmprintdatabase}.

\bibitem{ferrer2011GPDS}
M.~A. Ferrer, F.~Vargas, and A.~Morales, ``Bispectral contactless hand based
  biometric system,'' in \emph{CONATEL 2011}.\hskip 1em plus 0.5em minus
  0.4em\relax IEEE, 2011, pp. 1--6.

\bibitem{ZHANG2017199}
\BIBentryALTinterwordspacing
L.~Zhang, L.~Li, A.~Yang, Y.~Shen, and M.~Yang, ``Towards contactless palmprint
  recognition: A novel device, a new benchmark, and a collaborative
  representation based identification approach,'' \emph{Pattern Recognition},
  vol.~69, pp. 199 -- 212, 2017. [Online]. Available:
  \url{http://www.sciencedirect.com/science/article/pii/S0031320317301681}
\BIBentrySTDinterwordspacing

\bibitem{kumar2018HKPU_IITD_v3}
A.~Kumar, ``Toward more accurate matching of contactless palmprint images under
  less constrained environments,'' \emph{IEEE Transactions on Information
  Forensics and Security}, vol.~14, no.~1, pp. 34--47, 2018.

\bibitem{xiao2019palm_roi}
Q.~Xiao, J.~Lu, W.~Jia, and X.~Liu, ``Extracting palmprint roi from whole hand
  image using straight line clusters,'' \emph{IEEE Access}, 2019.

\bibitem{Aoyama2013}
S.~Aoyama, K.~Ito, T.~Aoki, and H.~Ota, ``{A Contactless Palmprint Recognition
  Algorithm for Mobile Phones},'' in \emph{Proceedings of International
  Workshop on Advanced Image Technology 2013}, Nagoya, 2013.

\bibitem{Kim2015}
\BIBentryALTinterwordspacing
J.~S. Kim, G.~Li, B.~Son, and J.~Kim, ``{An empirical study of palmprint
  recognition for mobile phones},'' \emph{IEEE Transactions on Consumer
  Electronics}, vol.~61, no.~3, pp. 311--319, aug 2015. [Online]. Available:
  \url{http://ieeexplore.ieee.org/lpdocs/epic03/wrapper.htm?arnumber=7298090}
\BIBentrySTDinterwordspacing

\bibitem{tiwari2016orb_pp}
K.~Tiwari, C.~J. Hwang, and P.~Gupta, ``A palmprint based recognition system
  for smartphone,'' in \emph{2016 Future Technologies Conference (FTC)}.\hskip
  1em plus 0.5em minus 0.4em\relax IEEE, 2016, pp. 577--586.

\bibitem{izadpana2019novel_mobile_datasets}
M.~Izadpanahkakhk, A.~Uncini, S.~H. Zahiri, M.~T. Gorjikolaie, and S.~M.
  Razavi, ``Novel mobile palmprint databases for biometric authentication,''
  \emph{International Journal of Grid and Utility Computing}, 2019.

\bibitem{Choras2012}
M.~Chora{\'{s}} and R.~Kozik, ``{Contactless palmprint and knuckle biometrics
  for mobile devices},'' \emph{Pattern Analysis and Applications}, vol.~15,
  no.~1, pp. 73--85, 2012.

\bibitem{Jia2012}
W.~Jia, R.~X. Hu, J.~Gui, Y.~Zhao, and X.~M. Ren, ``{Palmprint Recognition
  Across Different Devices},'' \emph{Sensors (Switzerland)}, vol.~12, no.~6,
  pp. 7938--7964, 2012.

\bibitem{afifi201711k}
M.~Afifi, ``11k hands: gender recognition and biometric identification using a
  large dataset of hand images,'' \emph{Multimedia Tools and Applications}, pp.
  1--20, 2017.

\bibitem{nuigpalmII2020dataset}
``Nuig-palm2 dataset of hand images,'' February 2020,
  {https://github.com/AdrianUng/NUIG-Palm2-palmprint-database}.

\bibitem{shao2019efficient}
H.~Shao, D.~Zhong, and X.~Du, ``Deep palmprint recognition via distilled
  hashing coding,'' in \emph{Proceedings of the IEEE Conference on Computer
  Vision and Pattern Recognition Workshops}, 2019, pp. 0--0.

\bibitem{zhang2019palm_roi}
Y.~Zhang, L.~Zhang, X.~Liu, S.~Zhao, Y.~Shen, and Y.~Yang, ``Pay by showing
  your palm: A study of palmprint verification on mobile platforms,'' in
  \emph{2019 IEEE International Conference on Multimedia and Expo
  (ICME)}.\hskip 1em plus 0.5em minus 0.4em\relax IEEE, 2019, pp. 862--867.

\bibitem{matkowski2019palmprint}
W.~M. Matkowski, T.~Chai, and A.~W.~K. Kong, ``Palmprint recognition in
  uncontrolled and uncooperative environment,'' \emph{IEEE Transactions on
  Information Forensics and Security}, 2019.

\bibitem{Wei2016}
S.~E. Wei, V.~Ramakrishna, T.~Kanade, and Y.~Sheikh, ``{Convolutional pose
  machines},'' \emph{Proceedings of the IEEE Computer Society Conference on
  Computer Vision and Pattern Recognition}, vol. 2016-Decem, pp. 4724--4732,
  2016.

\bibitem{otsu1979threshold}
N.~Otsu, ``A threshold selection method from gray-level histograms,''
  \emph{IEEE transactions on systems, man, and cybernetics}, vol.~9, no.~1, pp.
  62--66, 1979.

\bibitem{leng2014palm_roi}
L.~Leng, G.~Liu, M.~Li, M.~K. Khan, and A.~M. Al-Khouri, ``Logical conjunction
  of triple-perpendicular-directional translation residual for contactless
  palmprint preprocessing,'' in \emph{2014 11th International Conference on
  Information Technology: New Generations}.\hskip 1em plus 0.5em minus
  0.4em\relax IEEE, 2014, pp. 523--528.

\bibitem{zhang2003online}
D.~D. Zhang, W.~Kong, J.~You, and M.~Wong, ``Online palmprint identification,''
  \emph{IEEE Transactions on pattern analysis and machine intelligence}, 2003.

\bibitem{zhou2011palm_roi}
Y.~Zhou and A.~Kumar, ``Human identification using palm-vein images,''
  \emph{IEEE transactions on information forensics and security}, vol.~6,
  no.~4, pp. 1259--1274, 2011.

\bibitem{hao2008palm_roi}
Y.~Hao, Z.~Sun, T.~Tan, and C.~Ren, ``Multispectral palm image fusion for
  accurate contact-free palmprint recognition,'' in \emph{2008 15th IEEE
  International Conference on Image Processing}.\hskip 1em plus 0.5em minus
  0.4em\relax IEEE, 2008, pp. 281--284.

\bibitem{badrinath2012palm_roi}
G.~Badrinath and P.~Gupta, ``Palmprint based recognition system using
  phase-difference information,'' \emph{Future Generation Computer Systems},
  vol.~28, no.~1, pp. 287--305, 2012.

\bibitem{tiwari2013palm_roi}
K.~Tiwari, D.~K. Arya, G.~Badrinath, and P.~Gupta, ``Designing palmprint based
  recognition system using local structure tensor and force field
  transformation for human identification,'' \emph{Neurocomputing}, vol. 116,
  pp. 222--230, 2013.

\bibitem{hammami2014palm_roi}
M.~Hammami, S.~B. Jemaa, and H.~Ben-Abdallah, ``Selection of discriminative
  sub-regions for palmprint recognition,'' \emph{Multimedia tools and
  applications}, vol.~68, no.~3, pp. 1023--1050, 2014.

\bibitem{ito2015palm_roi}
K.~Ito, T.~Sato, S.~Aoyama, S.~Sakai, S.~Yusa, and T.~Aoki, ``Palm region
  extraction for contactless palmprint recognition,'' in \emph{2015
  International Conference on Biometrics (ICB)}.\hskip 1em plus 0.5em minus
  0.4em\relax IEEE, 2015, pp. 334--340.

\bibitem{poinsot2009palm_roi}
A.~{Poinsot}, F.~{Yang}, and M.~{Paindavoine}, ``Small sample biometric
  recognition based on palmprint and face fusion,'' in \emph{2009 Fourth
  International Multi-Conference on Computing in the Global Information
  Technology}, Aug 2009, pp. 118--122.

\bibitem{chen2007palm_roi}
W.~{Chen}, Y.~{Chiang}, and Y.~{Chiu}, ``Biometric verification by fusing hand
  geometry and palmprint,'' in \emph{Third International Conference on
  Intelligent Information Hiding and Multimedia Signal Processing (IIH-MSP
  2007)}, vol.~2, Nov 2007, pp. 403--406.

\bibitem{charfi2016sift_svm_fusion}
N.~Charfi, H.~Trichili, A.~M. Alimi, and B.~Solaiman, ``Local invariant
  representation for multi-instance toucheless palmprint identification,'' in
  \emph{2016 IEEE International Conference on Systems, Man, and Cybernetics
  (SMC)}.\hskip 1em plus 0.5em minus 0.4em\relax IEEE, 2016, pp.
  003\,522--003\,527.

\bibitem{balwant2015online}
M.~K. Balwant, A.~Agarwal, and C.~Rao, ``Online touchless palmprint
  registration system in a dynamic environment,'' \emph{Procedia Computer
  Science}, vol.~54, pp. 799--808, 2015.

\bibitem{gohkahong2008palm_roi}
G.~K.~O. Michael, T.~Connie, and A.~T.~B. Jin, ``Touch-less palm print
  biometric system.'' in \emph{VISAPP (2)}, 2008, pp. 423--430.

\bibitem{Franzgrote2011}
M.~Franzgrote, C.~Borg, B.~J.~T. Ries, S.~Bussemaker, X.~Jiang, M.~Fieleser,
  and L.~Zhang, ``Palmprint verification on mobile phones using accelerated
  competitive code,'' in \emph{2011 International Conference on Hand-Based
  Biometrics}.\hskip 1em plus 0.5em minus 0.4em\relax IEEE, 2011, pp. 1--6.

\bibitem{morales2012palm_roi}
A.~{Morales}, M.~A. {Ferrer}, C.~M. {Travieso}, and J.~B. {Alonso},
  ``Multisampling approach applied to contactless hand biometrics,'' in
  \emph{2012 IEEE International Carnahan Conference on Security Technology
  (ICCST)}, Oct 2012, pp. 224--229.

\bibitem{chai2016palm_roi}
T.~Chai, S.~Wang, and D.~Sun, ``A palmprint roi extraction method for mobile
  devices in complex environment,'' in \emph{2016 IEEE 13th International
  Conference on Signal Processing (ICSP)}.\hskip 1em plus 0.5em minus
  0.4em\relax IEEE, 2016, pp. 1342--1346.

\bibitem{sun2017palm_roi}
X.~{Sun}, Q.~{Xu}, C.~{Wang}, W.~{Dong}, and Z.~{Zhu}, ``Roi extraction for
  online touchless palm vein based on concavity analysis,'' in \emph{2017 32nd
  Youth Academic Annual Conference of Chinese Association of Automation (YAC)},
  May 2017, pp. 1123--1126.

\bibitem{khan2011contourcode}
Z.~Khan, A.~Mian, and Y.~Hu, ``Contour code: Robust and efficient multispectral
  palmprint encoding for human recognition,'' in \emph{2011 International
  Conference on Computer Vision}.\hskip 1em plus 0.5em minus 0.4em\relax IEEE,
  2011, pp. 1935--1942.

\bibitem{han2007palm_roi}
Y.~Han, Z.~Sun, F.~Wang, and T.~Tan, ``Palmprint recognition under
  unconstrained scenes,'' in \emph{Asian Conference on Computer Vision}.\hskip
  1em plus 0.5em minus 0.4em\relax Springer, 2007, pp. 1--11.

\bibitem{liang2019palm_roi}
X.~Liang, D.~Zhang, G.~Lu, Z.~Guo, and N.~Luo, ``A novel multicamera system for
  high-speed touchless palm recognition,'' \emph{IEEE Transactions on Systems,
  Man, and Cybernetics: Systems}, 2019.

\bibitem{yoruk2006hand_shape_recog}
E.~{Yoruk}, E.~{Konukoglu}, B.~{Sankur}, and J.~{Darbon}, ``Shape-based hand
  recognition,'' \emph{IEEE Transactions on Image Processing}, vol.~15, no.~7,
  pp. 1803--1815, July 2006.

\bibitem{jia2008rloc}
W.~Jia, D.-S. Huang, and D.~Zhang, ``Palmprint verification based on robust
  line orientation code,'' \emph{Pattern Recognition}, vol.~41, no.~5, pp.
  1504--1513, 2008.

\bibitem{shang2012pp_harris_roi}
L.~Shang, J.~Chen, P.-G. Su, and Y.~Zhou, ``Roi extraction of palmprint images
  using modified harris corner point detection algorithm,'' in
  \emph{International Conference on Intelligent Computing}.\hskip 1em plus
  0.5em minus 0.4em\relax Springer, 2012, pp. 479--486.

\bibitem{harris1988combined}
C.~G. Harris, M.~Stephens \emph{et~al.}, ``A combined corner and edge
  detector.'' in \emph{Alvey vision conference}, vol.~15, no.~50.\hskip 1em
  plus 0.5em minus 0.4em\relax Citeseer, 1988, pp. 10--5244.

\bibitem{javidnia2015istas}
H.~Javidnia, A.~Ungureanu, and P.~Corcoran, ``Palm-print recognition for
  authentication on smartphones,'' in \emph{2015 IEEE International Symposium
  on Technology and Society (ISTAS)}.\hskip 1em plus 0.5em minus 0.4em\relax
  IEEE, 2015, pp. 1--5.

\bibitem{Albiol2001}
\BIBentryALTinterwordspacing
A.~Albiol, L.~Torres, and E.~Delp, ``{Optimum color spaces for skin
  detection},'' in \emph{Proceedings 2001 International Conference on Image
  Processing (Cat. No.01CH37205)}, vol.~1, no.~xL.\hskip 1em plus 0.5em minus
  0.4em\relax IEEE, pp. 122--124. [Online]. Available:
  \url{http://ieeexplore.ieee.org/document/958968/}
\BIBentrySTDinterwordspacing

\bibitem{doublet2006contact}
J.~Doublet, O.~Lepetit, and M.~Revenu, ``Contact less hand recognition using
  shape and texture features,'' in \emph{Signal Processing, 2006 8th
  International Conference on}, vol.~3.\hskip 1em plus 0.5em minus 0.4em\relax
  IEEE, 2006.

\bibitem{Aykut2015}
M.~Aykut and M.~Ekinci, ``{Developing a contactless palmprint authentication
  system by introducing a novel ROI extraction method},'' \emph{Image and
  Vision Computing}, vol.~40, pp. 65--74, 2015.

\bibitem{kazemi2014kd_tree_regressor}
V.~Kazemi and J.~Sullivan, ``One millisecond face alignment with an ensemble of
  regression trees,'' in \emph{Proceedings of the IEEE conference on computer
  vision and pattern recognition}, 2014, pp. 1867--1874.

\bibitem{Bao2017}
X.~Bao and Z.~Guo, ``{Extracting region of interest for palmprint by
  convolutional neural networks},'' \emph{2016 6th International Conference on
  Image Processing Theory, Tools and Applications, IPTA 2016}, no. iii, 2017.

\bibitem{Izadpanahkakhk2018}
\BIBentryALTinterwordspacing
M.~Izadpanahkakhk, S.~Razavi, M.~Taghipour-Gorjikolaie, S.~Zahiri, and
  A.~Uncini, ``{Deep Region of Interest and Feature Extraction Models for
  Palmprint Verification Using Convolutional Neural Networks Transfer
  Learning},'' \emph{Applied Sciences}, vol.~8, no.~7, p. 1210, jul 2018.
  [Online]. Available: \url{http://www.mdpi.com/2076-3417/8/7/1210}
\BIBentrySTDinterwordspacing

\bibitem{chatfield2014return}
K.~Chatfield, K.~Simonyan, A.~Vedaldi, and A.~Zisserman, ``Return of the devil
  in the details: Delving deep into convolutional nets,'' \emph{arXiv preprint
  arXiv:1405.3531}, 2014.

\bibitem{jaswal2018deeppalm_roi}
G.~Jaswal, A.~Kaul, R.~Nath, and A.~Nigam, ``Deeppalm-a unified framework for
  personal human authentication,'' in \emph{2018 International Conference on
  Signal Processing and Communications (SPCOM)}.\hskip 1em plus 0.5em minus
  0.4em\relax IEEE, 2018, pp. 322--326.

\bibitem{fasterrcnn2015faster}
S.~Ren, K.~He, R.~Girshick, and J.~Sun, ``Faster r-cnn: Towards real-time
  object detection with region proposal networks,'' in \emph{Advances in neural
  information processing systems}, 2015, pp. 91--99.

\bibitem{liu2020shifted_triplet_loss}
Y.~Liu and A.~Kumar, ``Contactless palmprint identification using deeply
  learned residual features,'' \emph{IEEE Transactions on Biometrics, Behavior,
  and Identity Science}, 2020.

\bibitem{girshick2015fastRCNN}
R.~Girshick, ``Fast r-cnn,'' in \emph{Proceedings of the IEEE international
  conference on computer vision}, 2015, pp. 1440--1448.

\bibitem{jaderberg2015spatialTransf}
M.~Jaderberg, K.~Simonyan, A.~Zisserman \emph{et~al.}, ``Spatial transformer
  networks,'' in \emph{Advances in neural information processing systems},
  2015, pp. 2017--2025.

\bibitem{VGG16_initial}
K.~Simonyan and A.~Zisserman, ``Very deep convolutional networks for
  large-scale image recognition,'' \emph{CoRR}, vol. abs/1409.1556, 2014.

\bibitem{leng2018palm_roi}
L.~Leng, F.~Gao, Q.~Chen, and C.~Kim, ``Palmprint recognition system on mobile
  devices with double-line-single-point assistance,'' \emph{Personal and
  Ubiquitous Computing}, vol.~22, no.~1, pp. 93--104, 2018.

\bibitem{minaee2016palmprint}
S.~Minaee and Y.~Wang, ``Palmprint recognition using deep scattering
  convolutional network,'' \emph{arXiv preprint arXiv:1603.09027}, 2016.

\bibitem{meraoumia2016palm_pcanet}
A.~Meraoumia, F.~Kadri, H.~Bendjenna, S.~Chitroub, and A.~Bouridane,
  ``Improving biometric identification performance using pcanet deep learning
  and multispectral palmprint,'' in \emph{Biometric Security and
  Privacy}.\hskip 1em plus 0.5em minus 0.4em\relax Springer, 2016, pp. 51--69.

\bibitem{zhang2012palm_review}
D.~Zhang, W.~Zuo, and F.~Yue, ``A comparative study of palmprint recognition
  algorithms,'' \emph{ACM computing surveys (CSUR)}, vol.~44, no.~1, p.~2,
  2012.

\bibitem{kong2009palmprint_survey}
A.~Kong, D.~Zhang, and M.~Kamel, ``A survey of palmprint recognition,''
  \emph{pattern recognition}, vol.~42, no.~7, pp. 1408--1418, 2009.

\bibitem{dewangan2012palm_survey}
D.~P. Dewangan and A.~Pandey, ``A survey on security in palmprint recognitio: a
  biometric trait,'' \emph{Int. J. Adv. Res. Comput. Eng. Technol.(IJARCET)},
  vol.~1, p. 347, 2012.

\bibitem{jia2017complete_code}
W.~Jia, B.~Zhang, J.~Lu, Y.~Zhu, Y.~Zhao, W.~Zuo, and H.~Ling, ``Palmprint
  recognition based on complete direction representation,'' \emph{IEEE
  Transactions on Image Processing}, vol.~26, no.~9, pp. 4483--4498, 2017.

\bibitem{chen2010SAX}
J.~Chen, Y.-S. Moon, M.-F. Wong, and G.~Su, ``Palmprint authentication using a
  symbolic representation of images,'' \emph{Image and Vision Computing},
  vol.~28, no.~3, pp. 343--351, 2010.

\bibitem{raghavendra2014BSIF_palm}
R.~Raghavendra and C.~Busch, ``Robust palmprint verification using sparse
  representation of binarized statistical features: a comprehensive study,'' in
  \emph{Proceedings of the 2nd ACM workshop on Information hiding and
  multimedia security}.\hskip 1em plus 0.5em minus 0.4em\relax ACM, 2014, pp.
  181--185.

\bibitem{jia2013HOL}
W.~Jia, R.-X. Hu, Y.-K. Lei, Y.~Zhao, and J.~Gui, ``Histogram of oriented lines
  for palmprint recognition,'' \emph{IEEE Transactions on systems, man, and
  cybernetics: systems}, vol.~44, no.~3, pp. 385--395, 2013.

\bibitem{dalal2005HOG}
N.~Dalal and B.~Triggs, ``Histograms of oriented gradients for human
  detection,'' 2005.

\bibitem{jabid2010LDP}
T.~Jabid, M.~H. Kabir, and O.~Chae, ``Robust facial expression recognition
  based on local directional pattern,'' \emph{ETRI journal}, vol.~32, no.~5,
  pp. 784--794, 2010.

\bibitem{zheng2016don}
Q.~Zheng, A.~Kumar, and G.~Pan, ``A 3d feature descriptor recovered from a
  single 2d palmprint image,'' \emph{IEEE transactions on pattern analysis and
  machine intelligence}, vol.~38, no.~6, pp. 1272--1279, 2016.

\bibitem{li2017palmprint}
G.~Li and J.~Kim, ``Palmprint recognition with local micro-structure tetra
  pattern,'' \emph{Pattern Recognition}, vol.~61, pp. 29--46, 2017.

\bibitem{murala2012local_tetra_patterns}
S.~Murala, R.~Maheshwari, and R.~Balasubramanian, ``Local tetra patterns: a new
  feature descriptor for content-based image retrieval,'' \emph{IEEE
  transactions on image processing}, vol.~21, no.~5, pp. 2874--2886, 2012.

\bibitem{kong2004compcode}
A.-K. Kong and D.~Zhang, ``Competitive coding scheme for palmprint
  verification,'' in \emph{Pattern Recognition, 2004. ICPR 2004. Proceedings of
  the 17th International Conference on}, vol.~1.\hskip 1em plus 0.5em minus
  0.4em\relax IEEE, 2004, pp. 520--523.

\bibitem{wang2006palmprint}
X.~Wang, H.~Gong, H.~Zhang, B.~Li, and Z.~Zhuang, ``Palmprint identification
  using boosting local binary pattern,'' in \emph{18th International Conference
  on Pattern Recognition (ICPR'06)}, vol.~3.\hskip 1em plus 0.5em minus
  0.4em\relax IEEE, 2006, pp. 503--506.

\bibitem{ojala2002lbp}
T.~Ojala, M.~Pietik{\"a}inen, and T.~M{\"a}enp{\"a}{\"a}, ``Multiresolution
  gray-scale and rotation invariant texture classification with local binary
  patterns,'' \emph{IEEE Transactions on Pattern Analysis \& Machine
  Intelligence}, no.~7, pp. 971--987, 2002.

\bibitem{luo2016LLDP}
Y.-T. Luo, L.-Y. Zhao, B.~Zhang, W.~Jia, F.~Xue, J.-T. Lu, Y.-H. Zhu, and B.-Q.
  Xu, ``Local line directional pattern for palmprint recognition,''
  \emph{Pattern Recognition}, vol.~50, pp. 26--44, 2016.

\bibitem{sun2005olof}
Z.~Sun, T.~Tan, Y.~Wang, and S.~Z. Li, ``Ordinal palmprint represention for
  personal identification [represention read representation],'' in \emph{2005
  IEEE Computer Society Conference on Computer Vision and Pattern Recognition
  (CVPR'05)}, vol.~1.\hskip 1em plus 0.5em minus 0.4em\relax IEEE, 2005, pp.
  279--284.

\bibitem{wu2006dog_code}
X.~Wu, K.~Wang, and D.~Zhang, ``Palmprint texture analysis using derivative of
  gaussian filters,'' in \emph{2006 International Conference on Computational
  Intelligence and Security}, vol.~1.\hskip 1em plus 0.5em minus 0.4em\relax
  IEEE, 2006, pp. 751--754.

\bibitem{guo2009BOCV}
Z.~Guo, D.~Zhang, L.~Zhang, and W.~Zuo, ``Palmprint verification using binary
  orientation co-occurrence vector,'' \emph{Pattern Recognition Letters},
  vol.~30, no.~13, pp. 1219--1227, 2009.

\bibitem{zhang2009hkpu_ms}
D.~Zhang, Z.~Guo, G.~Lu, L.~Zhang, and W.~Zuo, ``An online system of
  multispectral palmprint verification,'' \emph{IEEE transactions on
  instrumentation and measurement}, vol.~59, no.~2, pp. 480--490, 2009.

\bibitem{hao2008casia_ms}
Y.~Hao, Z.~Sun, T.~Tan, and C.~Ren, ``Multispectral palm image fusion for
  accurate contact-free palmprint recognition,'' in \emph{2008 15th IEEE
  International Conference on Image Processing}.\hskip 1em plus 0.5em minus
  0.4em\relax IEEE, 2008, pp. 281--284.

\bibitem{zhang2012EBOCV}
L.~Zhang, H.~Li, and J.~Niu, ``Fragile bits in palmprint recognition,''
  \emph{IEEE Signal processing letters}, vol.~19, no.~10, pp. 663--666, 2012.

\bibitem{fei2016DOC}
L.~Fei, Y.~Xu, W.~Tang, and D.~Zhang, ``Double-orientation code and nonlinear
  matching scheme for palmprint recognition,'' \emph{Pattern Recognition},
  vol.~49, pp. 89--101, 2016.

\bibitem{zheng2016fast_compcode}
Q.~Zheng, A.~Kumar, and G.~Pan, ``Suspecting less and doing better: New
  insights on palmprint identification for faster and more accurate matching,''
  \emph{IEEE Transactions on Information Forensics and Security}, vol.~11,
  no.~3, pp. 633--641, 2016.

\bibitem{fei2016half}
L.~Fei, Y.~Xu, and D.~Zhang, ``Half-orientation extraction of palmprint
  features,'' \emph{Pattern Recognition Letters}, vol.~69, pp. 35--41, 2016.

\bibitem{tabejamaat2017concavity_banana}
M.~Tabejamaat and A.~Mousavi, ``Concavity-orientation coding for palmprint
  recognition,'' \emph{Multimedia Tools and Applications}, vol.~76, no.~7, pp.
  9387--9403, 2017.

\bibitem{fei2016neighb_direction_indicator}
L.~Fei, B.~Zhang, Y.~Xu, and L.~Yan, ``Palmprint recognition using neighboring
  direction indicator,'' \emph{IEEE Transactions on Human-Machine Systems},
  vol.~46, no.~6, pp. 787--798, 2016.

\bibitem{fei2016LMDP}
L.~Fei, J.~Wen, Z.~Zhang, K.~Yan, and Z.~Zhong, ``Local multiple directional
  pattern of palmprint image,'' in \emph{2016 23rd International Conference on
  Pattern Recognition (ICPR)}.\hskip 1em plus 0.5em minus 0.4em\relax IEEE,
  2016, pp. 3013--3018.

\bibitem{xu2016robust_compcode}
Y.~Xu, L.~Fei, J.~Wen, and D.~Zhang, ``Discriminative and robust competitive
  code for palmprint recognition,'' \emph{IEEE Transactions on Systems, Man,
  and Cybernetics: Systems}, vol.~48, no.~2, pp. 232--241, 2016.

\bibitem{chen2008sift_pp}
J.~Chen and Y.-S. Moon, ``Using sift features in palmprint authentication,'' in
  \emph{2008 19th International Conference on Pattern Recognition}.\hskip 1em
  plus 0.5em minus 0.4em\relax IEEE, 2008, pp. 1--4.

\bibitem{Morales2011}
\BIBentryALTinterwordspacing
A.~Morales, M.~Ferrer, and A.~Kumar, ``{Towards contactless palmprint
  authentication},'' \emph{IET Computer Vision}, vol.~5, no.~6, p. 407, 2011.
  [Online]. Available:
  \url{http://digital-library.theiet.org/content/journals/10.1049/iet-cvi.2010.0191}
\BIBentrySTDinterwordspacing

\bibitem{Zhao2013sift_iransac}
\BIBentryALTinterwordspacing
Q.~Zhao, X.~Wu, and W.~Bu, ``{Contactless palmprint verification based on SIFT
  and iterative RANSAC},'' in \emph{2013 IEEE International Conference on Image
  Processing}.\hskip 1em plus 0.5em minus 0.4em\relax IEEE, sep 2013, pp.
  4186--4189. [Online]. Available:
  \url{http://ieeexplore.ieee.org/lpdocs/epic03/wrapper.htm?arnumber=6738862}
\BIBentrySTDinterwordspacing

\bibitem{kang2014mod_sift}
W.~Kang, Y.~Liu, Q.~Wu, and X.~Yue, ``Contact-free palm-vein recognition based
  on local invariant features,'' \emph{PloS one}, vol.~9, no.~5, p. e97548,
  2014.

\bibitem{hollingsworth2008iris_code_fragile_bits}
K.~P. Hollingsworth, K.~W. Bowyer, and P.~J. Flynn, ``The best bits in an iris
  code,'' \emph{IEEE Transactions on Pattern Analysis and Machine
  Intelligence}, vol.~31, no.~6, pp. 964--973, 2008.

\bibitem{peters1997banana_wavelet}
G.~Peters, N.~Kr{\"u}ger, and C.~Von Der~Malsburg, ``Learning object
  representations by clustering banana wavelet responses,'' \emph{Proceedings
  of the 1st STIPR}, pp. 113--118, 1997.

\bibitem{zhong2013ELDP}
F.~Zhong and J.~Zhang, ``Face recognition with enhanced local directional
  patterns,'' \emph{Neurocomputing}, vol. 119, pp. 375--384, 2013.

\bibitem{rivera2012LDN}
A.~R. Rivera, J.~R. Castillo, and O.~O. Chae, ``Local directional number
  pattern for face analysis: Face and expression recognition,'' \emph{IEEE
  transactions on image processing}, vol.~22, no.~5, pp. 1740--1752, 2012.

\bibitem{fei2018complete_binary_repr}
L.~Fei, G.~Lu, W.~Jia, J.~Wen, and D.~Zhang, ``Complete binary representation
  for 3-d palmprint recognition,'' \emph{IEEE Transactions on Instrumentation
  and Measurement}, vol.~67, no.~12, pp. 2761--2771, 2018.

\bibitem{iitsuka2008BLPOC_palm}
S.~Iitsuka, K.~Ito, and T.~Aoki, ``A practical palmprint recognition algorithm
  using phase information,'' in \emph{2008 19th International Conference on
  Pattern Recognition}.\hskip 1em plus 0.5em minus 0.4em\relax IEEE, 2008, pp.
  1--4.

\bibitem{Lowe2004SIFT}
\BIBentryALTinterwordspacing
D.~G. Lowe, ``{Distinctive Image Features from Scale-Invariant Keypoints},''
  \emph{International Journal of Computer Vision}, vol.~60, no.~2, pp. 91--110,
  nov 2004. [Online]. Available:
  \url{http://link.springer.com/10.1023/B:VISI.0000029664.99615.94}
\BIBentrySTDinterwordspacing

\bibitem{charfi2014bosph_sift}
N.~Charfi, H.~Trichili, A.~M. Alimi, and B.~Solaiman, ``Novel hand biometric
  system using invariant descriptors,'' in \emph{2014 6th International
  Conference of Soft Computing and Pattern Recognition (SoCPaR)}.\hskip 1em
  plus 0.5em minus 0.4em\relax IEEE, 2014, pp. 261--266.

\bibitem{rublee2011ORB}
E.~Rublee, V.~Rabaud, K.~Konolige, and G.~R. Bradski, ``Orb: An efficient
  alternative to sift or surf.'' in \emph{ICCV}, vol.~11, no.~1.\hskip 1em plus
  0.5em minus 0.4em\relax Citeseer, 2011, p.~2.

\bibitem{srinivas2009palmprint_surf}
B.~G. Srinivas and P.~Gupta, ``Palmprint based verification system using surf
  features,'' in \emph{International Conference on Contemporary
  Computing}.\hskip 1em plus 0.5em minus 0.4em\relax Springer, 2009, pp.
  250--262.

\bibitem{Bay2008}
\BIBentryALTinterwordspacing
H.~Bay, A.~Ess, T.~Tuytelaars, and L.~{Van Gool}, ``{Speeded-Up Robust Features
  (SURF)},'' \emph{Computer Vision and Image Understanding}, vol. 110, no.~3,
  pp. 346--359, jun 2008. [Online]. Available:
  \url{http://linkinghub.elsevier.com/retrieve/pii/S1077314207001555}
\BIBentrySTDinterwordspacing

\bibitem{dian2016palmprint_cnn}
L.~Dian and S.~Dongmei, ``Contactless palmprint recognition based on
  convolutional neural network,'' in \emph{2016 IEEE 13th International
  Conference on Signal Processing (ICSP)}.\hskip 1em plus 0.5em minus
  0.4em\relax IEEE, 2016, pp. 1363--1367.

\bibitem{tarawneh2018}
A.~S. Tarawneh, D.~Chetverikov, and A.~B. Hassanat, ``Pilot comparative study
  of different deep features for palmprint identification in low-quality
  images,'' \emph{arXiv preprint arXiv:1804.04602}, 2018.

\bibitem{hassanat2015mohi}
A.~Hassanat, M.~Al-Awadi, E.~Btoush, A.~Al-Btoush, G.~Altarawneh \emph{et~al.},
  ``New mobile phone and webcam hand images databases for personal
  authentication and identification,'' \emph{Procedia Manufacturing}, vol.~3,
  pp. 4060--4067, 2015.

\bibitem{ramachandra2018icb}
R.~Ramachandra, K.~B. Raja, S.~Venkatesh, S.~Hegde, S.~D. Dandappanavar, and
  C.~Busch, ``Verifying the newborns without infection risks using contactless
  palmprints,'' in \emph{2018 International Conference on Biometrics
  (ICB)}.\hskip 1em plus 0.5em minus 0.4em\relax IEEE, 2018, pp. 209--216.

\bibitem{genovese2019palmnet}
A.~Genovese, V.~Piuri, K.~N. Plataniotis, and F.~Scotti, ``Palmnet: Gabor-pca
  convolutional networks for touchless palmprint recognition,'' \emph{IEEE
  Transactions on Information Forensics and Security}, 2019.

\bibitem{zhang2015collaborative_3d_pp}
L.~Zhang, Y.~Shen, H.~Li, and J.~Lu, ``3d palmprint identification using
  block-wise features and collaborative representation,'' \emph{IEEE
  transactions on pattern analysis and machine intelligence}, vol.~37, no.~8,
  pp. 1730--1736, 2015.

\bibitem{jalali2015cnn}
A.~Jalali, R.~Mallipeddi, and M.~Lee, ``Deformation invariant and contactless
  palmprint recognition using convolutional neural network,'' in
  \emph{Proceedings of the 3rd International Conference on Human-Agent
  Interaction}.\hskip 1em plus 0.5em minus 0.4em\relax ACM, 2015, pp. 209--212.

\bibitem{zhang2018palmprint}
L.~Zhang, Z.~Cheng, Y.~Shen, and D.~Wang, ``Palmprint and palmvein recognition
  based on dcnn and a new large-scale contactless palmvein dataset,''
  \emph{Symmetry}, vol.~10, no.~4, p.~78, 2018.

\bibitem{fei2018pp_feat_extr_review}
L.~Fei, G.~Lu, W.~Jia, S.~Teng, and D.~Zhang, ``Feature extraction methods for
  palmprint recognition: A survey and evaluation,'' \emph{IEEE Transactions on
  Systems, Man, and Cybernetics: Systems}, vol.~49, no.~2, pp. 346--363, 2018.

\bibitem{zhao2019joint_cnn_palm}
S.~Zhao, B.~Zhang, and C.~P. Chen, ``Joint deep convolutional feature
  representation for hyperspectral palmprint recognition,'' \emph{Information
  Sciences}, vol. 489, pp. 167--181, 2019.

\bibitem{geladi1986PLS}
P.~Geladi and B.~R. Kowalski, ``Partial least-squares regression: a tutorial,''
  \emph{Analytica chimica acta}, vol. 185, pp. 1--17, 1986.

\bibitem{svoboda2016palmprint}
J.~Svoboda, J.~Masci, and M.~M. Bronstein, ``Palmprint recognition via
  discriminative index learning,'' in \emph{2016 23rd International Conference
  on Pattern Recognition (ICPR)}.\hskip 1em plus 0.5em minus 0.4em\relax IEEE,
  2016, pp. 4232--4237.

\bibitem{zhong2018siamese}
D.~Zhong, Y.~Yang, and X.~Du, ``Palmprint recognition using siamese network,''
  in \emph{Chinese Conference on Biometric Recognition}.\hskip 1em plus 0.5em
  minus 0.4em\relax Springer, 2018, pp. 48--55.

\bibitem{hinton2015distilling}
G.~Hinton, O.~Vinyals, and J.~Dean, ``Distilling the knowledge in a neural
  network,'' \emph{arXiv preprint arXiv:1503.02531}, 2015.

\bibitem{shao2019palmgan}
H.~Shao, D.~Zhong, and Y.~Li, ``Palmgan for cross-domain palmprint
  recognition,'' in \emph{2019 IEEE International Conference on Multimedia and
  Expo (ICME)}.\hskip 1em plus 0.5em minus 0.4em\relax IEEE, 2019, pp.
  1390--1395.

\bibitem{du2019low_shot}
X.~Du, D.~Zhong, and P.~Li, ``Low-shot palmprint recognition based on
  meta-siamese network,'' in \emph{2019 IEEE International Conference on
  Multimedia and Expo (ICME)}.\hskip 1em plus 0.5em minus 0.4em\relax IEEE,
  2019, pp. 79--84.

\bibitem{krizhevsky2012alexnet}
A.~Krizhevsky, I.~Sutskever, and G.~E. Hinton, ``Imagenet classification with
  deep convolutional neural networks,'' in \emph{Advances in neural information
  processing systems}, 2012, pp. 1097--1105.

\bibitem{bruna2013scattering_networks}
J.~Bruna and S.~Mallat, ``Invariant scattering convolution networks,''
  \emph{IEEE transactions on pattern analysis and machine intelligence},
  vol.~35, no.~8, pp. 1872--1886, 2013.

\bibitem{chan2015pcanet}
T.-H. Chan, K.~Jia, S.~Gao, J.~Lu, Z.~Zeng, and Y.~Ma, ``Pcanet: A simple deep
  learning baseline for image classification?'' \emph{IEEE transactions on
  image processing}, vol.~24, no.~12, pp. 5017--5032, 2015.

\bibitem{hadsell2006contrastive_loss}
R.~Hadsell, S.~Chopra, and Y.~LeCun, ``Dimensionality reduction by learning an
  invariant mapping,'' in \emph{2006 IEEE Computer Society Conference on
  Computer Vision and Pattern Recognition (CVPR'06)}, vol.~2.\hskip 1em plus
  0.5em minus 0.4em\relax IEEE, 2006, pp. 1735--1742.

\bibitem{wen2016discriminative}
Y.~Wen, K.~Zhang, Z.~Li, and Y.~Qiao, ``A discriminative feature learning
  approach for deep face recognition,'' in \emph{European conference on
  computer vision}.\hskip 1em plus 0.5em minus 0.4em\relax Springer, 2016, pp.
  499--515.

\bibitem{schroff2015facenet}
F.~Schroff, D.~Kalenichenko, and J.~Philbin, ``Facenet: A unified embedding for
  face recognition and clustering,'' in \emph{Proceedings of the IEEE
  conference on computer vision and pattern recognition}, 2015, pp. 815--823.

\bibitem{howard2017mobilenets}
A.~G. Howard, M.~Zhu, B.~Chen, D.~Kalenichenko, W.~Wang, T.~Weyand,
  M.~Andreetto, and H.~Adam, ``Mobilenets: Efficient convolutional neural
  networks for mobile vision applications,'' \emph{arXiv preprint
  arXiv:1704.04861}, 2017.

\bibitem{shao2019neural_graphs}
H.~Shao and D.~Zhong, ``Few-shot palmprint recognition via graph neural
  networks,'' \emph{Electronics Letters}, 2019.

\bibitem{zhu2017cycleGAN}
J.-Y. Zhu, T.~Park, P.~Isola, and A.~A. Efros, ``Unpaired image-to-image
  translation using cycle-consistent adversarial networks,'' in
  \emph{Proceedings of the IEEE international conference on computer vision},
  2017, pp. 2223--2232.

\bibitem{kanhangad2010pose_invariant}
V.~Kanhangad, A.~Kumar, and D.~Zhang, ``Contactless and pose invariant
  biometric identification using hand surface,'' \emph{IEEE transactions on
  image processing}, vol.~20, no.~5, pp. 1415--1424, 2010.

\bibitem{Mueller2017}
\BIBentryALTinterwordspacing
F.~Mueller, F.~Bernard, O.~Sotnychenko, D.~Mehta, S.~Sridhar, D.~Casas, and
  C.~Theobalt, ``Ganerated hands for real-time 3d hand tracking from monocular
  rgb,'' in \emph{Proceedings of Computer Vision and Pattern Recognition
  ({CVPR})}, June 2018. [Online]. Available:
  \url{https://handtracker.mpi-inf.mpg.de/projects/GANeratedHands/}
\BIBentrySTDinterwordspacing

\bibitem{li2016ternary}
F.~Li, B.~Zhang, and B.~Liu, ``Ternary weight networks,'' \emph{arXiv preprint
  arXiv:1605.04711}, 2016.

\end{thebibliography}


%\end{IEEEbiography}

%\newpage



%\vfill


\end{document}